\documentclass[10pt]{article}
\usepackage[dvipsnames]{xcolor}
\usepackage[preprint]{tmlr}

\usepackage{graphicx} %
\usepackage[hidelinks]{hyperref}
\usepackage{url}
\usepackage{amsmath}
\usepackage{amssymb}
\usepackage{mathtools}
\usepackage{cancel}
\usepackage[capitalize]{cleveref}
\usepackage{bm}
\usepackage[small]{caption}
\usepackage{booktabs}
\usepackage{enumitem}
\usepackage[linesnumbered,ruled,vlined]{algorithm2e}
\usepackage{nicefrac}
\usepackage{comment}
\usepackage{siunitx}
\Crefname{algocf}{Algorithm}{Algorithms}

\SetCommentSty{mycommfont}
\SetKwInput{KwInput}{Input}                %
\SetKwInput{KwOutput}{Output}              %

\title{Introduction to Stochastic Differential Equations for Generative Machine Learning: A Variational Perspective}
\author{\name Ole Winther \email ole.winther@bio.ku.dk \\
        \addr Department of Biology, University of Copenhagen, 2200, Copenhagen, Denmark\\Department of Applied Mathematics and Computer Science, Technical University of Denmark, Kgs. Lyngby, Denmark\looseness-1
        \AND
        \name Paul Jeha \email pauje@dtu.dk \\
        \addr Department of Applied Mathematics and Computer Science, Technical University of Denmark, Kgs. Lyngby, Denmark\looseness-1
        \AND
        \name Sander Dieleman \email sedielem@deepmind.com \\ 
        \addr Google DeepMind, London, UK
        \AND
        \name Andriy Mnih \email andriy@deepmind.com \\
        \addr Google DeepMind, London, UK
        \AND
        \name Manfred Opper \email manfred.opper@tu-berlin.de \\
        \addr Technical University of Berlin (TU Berlin), Berlin, Germany
        \AND
        \name Andrea Dittadi \email andrea.dittadi@gmail.com \\
        \addr Technical University of Munich, Germany}

\date{June 2026}

\newcommand{\x}{\mathbf{x}}
\newcommand{\y}{\mathbf{y}}
\newcommand{\f}{\mathbf{f}}
\newcommand{\g}{\mathbf{g}}
\newcommand{\s}{\mathbf{s}}
\renewcommand{\v}{\mathbf{v}}
\newcommand{\A}{\mathbf{A}}
\newcommand{\D}{\mathbf{D}}
\newcommand{\F}{\mathbf{F}}
\newcommand{\G}{\mathbf{G}}
\newcommand{\W}{\mathbf{W}}
\newcommand{\I}{\mathbf{I}}
\newcommand{\X}{\mathbf{X}}

\newcommand{\0}{\mathbf{0}}

\newcommand{\alphab}{\bm{\alpha}}

\newcommand{\sigmab}{\bm{\sigma}}

\newcommand{\epsilonb}{\bm{\epsilon}}

\newcommand{\phib}{\bm{\phi}}

\newcommand{\thetab}{\boldsymbol{\theta}}
\newcommand{\zerob}{\mathbf{0}}

\newcommand{\diff}{\mathrm{d}}
\newcommand{\dx}{\diff \x}
\newcommand{\dt}{\diff t}

\newcommand{\KL}{D_{\mathrm{KL}}}
\newcommand{\Qpath}{\mathbb{Q}}
\newcommand{\E}{\mathbb{E}}
\newcommand{\R}{\mathbb{R}}
\newcommand{\N}{\mathbb{N}}

\DeclareMathOperator*{\argmin}{argmin}
\DeclareMathOperator{\SNR}{SNR}

\newcommand{\transp}{^{\mathrm{T}}}

\newcommand{\eqnum}[1]{\stackrel{(#1)}{=}}

\newcommand{\ODEELBO}{\mathcal{L}_{\mathrm{ELBO}}^{\mathrm{ODE}}}
\newcommand{\SDEELBO}{\mathcal{L}_{\mathrm{ELBO}}^{\mathrm{SDE}}}

\begin{document}

\maketitle

\begin{abstract}
    The use of ordinary and stochastic differential equations has led to substantial progress in generative machine learning with applications to, for example, image, video and biomolecule generation. This paper provides a self-contained and informal introduction to the differential equations, the probabilistic framework for using them in generative modeling and the Fokker--Planck equation that governs the temporal evolution of the marginal distribution of the stochastic variables of the differential equations. The variational lower bound on the log-likelihood (the evidence lower bound, ELBO) is derived and used as a general starting point for a discussion of diffusion models, score matching, and flow matching. All of these approaches may be viewed as specific parameterizations of the most general variational approach. A one-dimensional density modeling problem is used as a simple example to compare different parameterizations.
\end{abstract}

\section{Introduction}

In recent years, stochastic processes have played a fundamental role in generative machine learning, as exemplified by neural ordinary differential equations (Neural ODEs) \citep{chen2018neural}, score matching \citep{song2019generative,song2020denoising,song2020score,song2021maximum,song2023consistency}, diffusion models \citep{sohl2015deep,ho2020denoising,kingma2021variational}, flow matching \citep{lipman2022flow,liu2022flow,tong2023improving} and stochastic interpolants \citep{albergo2023stochastic}. These methodologies have led to applied breakthroughs in, for example, image and video generation and bio-molecule structure prediction and design \citep{ho2020denoising, ho2022video, hoogeboom2022equivariantdiffusionmoleculegeneration}. It is noteworthy that older work, such as \citet{archambeau2007gaussian,opper2007variational}, developed a similar methodology with applications in the time series setting rather than in generative modeling.

Most publications assume a relatively advanced level understanding of stochastic processes and probabilistic methods. The exposition typically starts from a statement of the generative model as a stochastic process written in terms of an ordinary or stochastic differential equation (ODE or SDE) with fundamental concepts, such as the Fokker-Planck equation, being applied without being formally derived. In addition, recent advances often present different perspectives on similar concepts, making it challenging to understand connections between them without a strong background in the basic methodology. On the other end of the spectrum, many good primers have appeared online that have strong intuitive explanations starting from simulations and heuristic objective functions, but lack connection to the probabilistic foundation.

This has motivated us to provide a self-contained introduction to the fundamentals of these methods and present methods from the machine learning literature in a unified way. In some instances, our variational perspective may differ slightly from the prevailing paradigm, with the goal of offering a fresh view on popular methods.

The approach we take here is to focus on \emph{how to} derive the results appearing in the machine learning literature with less focus on formality and mathematical rigor. Specifically, the master Fokker--Planck equation that governs the temporal evolution of the marginals distribution of the stochastic variables and the reverse-time SDE are often presented as fundamental results. 
The exposition is sought to be self-contained with minimal need for specialized background knowledge for example neither Itô calculus nor the Girsanov theorem will be used directly. Rather we will appeal to using Taylor expansions or start from finite differences equation and show without proving that the continuous formulation follows from taking the infinitesimal limit. The necessary assumptions (such as existence of derivatives to different orders or integrands vanishing at integration limits) will be introduced where they occur in the derivation rather than upfront, to avoid overloading the reader.

This paper is not intended as a review. Our aim is to provide a key to unlock the literature at the intersection of machine learning and stochastic processes. As a result, many recent methodological developments, such as more computationally efficient sampling techniques and constraints applied to stochastic processes to endow models with specific properties, are not covered. It is assumed that the reader has good understanding of calculus and probability and thus the paper will not be the ideal starting point for all readers. 

The main contributions of this paper are:
\begin{itemize}
    \item A self-contained presentation of probabilistic methodologies, including the derivation of the Fokker--Planck equation starting from the definition of the ordinary and stochastic differential equations (ODEs and SDEs), consolidating this content in one place rather than being scattered across textbooks, stochastic process literature, and appendices in machine learning papers. The next three sections are dedicated to this: ODEs and SDEs (Section \ref{sec: ODE and SDE}), generative models (Section \ref{sec: Generative models from differential equations}) and the Fokker-Planck equation (Section \ref{sec: Fokker--Planck}).
    \item A unified derivation of objectives used in generative modeling with ODEs and SDEs, including diffusion, score and flow matching, all presented within a general variational framework. Three sections are dedicated to this: training objective (Section \ref{sec: likelihood}), parameterization of the SDE objective (Section \ref{sec: drift from flow}) and diffusion, score and flow matching (Section \ref{sec: diffusion score flow}).
    \item A methodology for constructing analytically tractable processes that extend beyond those with Gaussian marginals. The motivation for pushing these limits is inspire researchers to go beyond how these methods are used today described in Section \ref{sec: drift from flow}.
\end{itemize}
Sections \ref{sec: experiments} and \ref{sec: discussion} contain a simple one-dimensional density modeling example comparison and the discussion.

\section{Ordinary and stochastic differential equations (ODEs and SDEs)} \label{sec: ODE and SDE}

We begin by introducing the fundamentals of differential equations. We will consider a stochastic variable $\x_t \in \R^d$ indexed by a continuous \emph{time} parameter $t \in [0,1]$.
The evolution of $\x_t$ is described in terms of infinitesimal changes, expressed using the differential notation $\diff\x_t$, which can be seen as the limit of $\Delta \x_t = \x_{t + \Delta t} - \x_t$ for $\Delta t \to 0$.
This limit is not formally defined on its own, but serves as an intuitive and compact representation of local dynamics. A rigorous understanding can be achieved using an integral formulation.

In this work, we will focus on Markov processes, where the system's infinitesimal evolution $\diff\x_t$ depends on the current time $t$ and the past through the current state $\x_t$ only. 
As a starting point, we consider the deterministic case, where the system evolves according to an \emph{ordinary differential equation} (ODE): $\frac{\diff \x_t}{\diff t} = \f(\x_t,t)$, where the \emph{drift} $\f : \R^d \times [0,1] \to \R^d$ specifies the instantaneous rate of change of the state. We will write the ODE as 
\begin{align}
    \diff\x_t &= \f(\x_t,t)\diff t \ . \mbox{\hspace{4.7cm}(ODE)}
    \label{eq:ODE}
\end{align}
This is an informal notation, not rigorously defined in the infinitesimal limit, but it allows  us to treat the deterministic and stochastic versions in a unified way.  
To model systems influenced by random fluctuations, we extend this formulation to \emph{stochastic differential equations} (SDEs), which include an additional stochastic term:
\begin{align}
    \diff\x_t & = \f(\x_t,t)\diff t+ \sigmab(\x_t,t) \diff\W_t \mbox{\hspace{2.5cm}(SDE)}
    \label{eq:SDE}
\end{align}
Here, the \emph{diffusion coefficient} $\sigmab: \R^d \times [0,1] \to \R^{d \times m}$ determines how noise enters and influences each component of the state, and $\diff\W_t$ denotes the infinitesimal stochastic increment of an $m$-dimensional \emph{Wiener process} (or Brownian motion). There are no restrictions on the choice of $\sigmab$ other than in one specific case (Section \ref{sec: latent SDE}), we will require that $\sigmab \sigmab\transp$ is full rank. 

A Wiener process $\W_t$ is a continuous-time stochastic process such that (1) $\W_0 = \zerob$ almost surely (i.e., with probability 1); (2) its increments are distributed as $\W_{t + \Delta t} - \W_t \sim \mathcal{N}(\zerob, \Delta t \, \I)$ for $\Delta t > 0$ and (3) is independent of past values $\{\W_s : s<t\}$; (4) its sample paths are almost surely continuous in $t$.
The term $\diff\W_t$ is thus a symbolic representation of an increment of the Wiener process over an infinitesimal time interval. It introduces a source of noise that is independent of the current system state. This independence can be defined as $p_t(\x_t, \diff\W_t) = p_t(\x_t) p(\diff\W_t)$, where $p_t(\x_t)$ is the \emph{marginal distribution of $\x_t$ at time $t$}. This leads to the so-called Itô SDE that we study in this paper. The alternative but equivalent Stratonovich formulation with dependent noise is sometimes more convenient \citep{gardiner2009stochastic,oksendal2013stochastic}. 

In practice, solutions to both ODEs and SDEs are often approximated using time discretization methods. Typically, time is discretized by partitioning the interval $[0,1]$ into $T$ subintervals of length $\Delta t = \nicefrac{1}{T}$. For stochastic systems, the most basic and widely used approach is the \emph{Euler--Maruyama} scheme, a stochastic analogue of the classical Euler method. The evolution of the process is approximated by
\begin{align}
    \x_{t+\Delta t} \approx \x_t + \f(\x_t, t) \;\Delta t + \sigmab(\x_t, t) \;\Delta\W_t \ ,
\end{align}
where $\Delta \W_t \sim\mathcal{N}(0,\Delta t \, \I)$.
This method provides a foundation for both numerical simulations and theoretical analysis of stochastic dynamics.
As $T \to \infty$, the discrete approximation converges to the marginal distribution and under some conditions to the full distribution of the continuous-time Wiener process \citep{platen1999introduction}, and $\Delta t, \Delta \W_t$ are replaced by the differentials $\diff t, \diff\W_t$.

\section{Generative differential equations}
\label{sec: Generative models from differential equations}

In this paper we will only consider differential equations for generative modeling.
We assume an observed (continuous or discrete) random variable $\y$ with an unknown true distribution $q(\y)$, and our goal is to learn a model distribution $p(\y)$ that approximates $q(\y)$. Here we follow the convention from the infinite data limit derivation of the score and flow matching methods to use $q(\y)$ for the data distribution, see \cref{app:score and flow}. Additionally, we aim to efficiently generate new samples $\y \sim p(\y)$.
The model specification consists of
\begin{enumerate}[label=(\roman*)]
    \item a latent process $\{\x_t\}_{t\in[0,1]}$ in $\R^d$, described by the ODE \eqref{eq:ODE} or the SDE \eqref{eq:SDE}, which starts from
    \item a simple \emph{prior} distribution $p_0(\x_0)$ at $t=0$, and 
    \item a distribution $p(\y|\x_1)$ conditioned on $\x_1$, typically called \emph{likelihood function}.
\end{enumerate}
Using a latent process to model $p(\y)$ is a powerful construction for flexible modeling. The latent process in general will provide a smooth non-linear transformation of the latent variable and thus is more expressive than the one-shot variational autoencoder approach of using a likelihood function $p(\y|\x)$ with a simple prior distribution $p(\x)$. 
To learn the generative model, we will assume access to a training set of $n$ independent and identically distributed (iid) samples $\{\y^{(1)},\ldots,\y^{(n)}\}$ from the data distribution $q(\y)$.

Formally, the prior and the latent process induce a distribution $p(\X)$ over the continuous-time paths $\X=\{\x_t\}_{t\in[0,1]}$. We note that the joint distribution $p(\X) = p(\x_{t_0}, \x_{t_1}, \ldots, \x_{t_T})$ only exists for discretizations of the process with a finite number of steps $0=t_0<\ldots<t_T=1$, and is not well-defined in the $T\to\infty$ limit.
Nevertheless, we will adopt this informal perspective as it makes the derivations more manageable and helps build an intuitive understanding. The \emph{likelihood} (the probability of the data point $\y$ under the model)
can now formally be written as a marginalization over the latent paths:
\begin{equation}
    \label{eq:likelihood}
    p(\y)=\int p(\y|\x_1) p(\X) \diff\X = \int p(\y|\x_1) p_1(\x_1) \diff\x_1 \ ,
\end{equation}
which shows that we can compute the (marginal) likelihood if we have access to the marginal distribution at $t=1$.

Computing the likelihood is in general analytically intractable, and (bridge, nested and tempered) Monte Carlo estimation is computationally expensive. The variational approach presented in \cref{sec: likelihood} provides a computationally efficient stochastic estimate of a lower bound to the log-likelihood, making it attractive for large-scale machine learning. The log-likelihood lower bound is commonly referred to as the evidence lower bound (ELBO).

Both the exact (\cref{eq:likelihood}) and approximate (\cref{sec: likelihood}) computation of the log-likelihood require integration over path distributions (path integrals). The path integral derivation is given in \cref{app: girsanov}. Alternatively, the marginal $p_1(\x_1)$ in \eqref{eq:likelihood} can be obtained as the solution of a differential equation for the marginal $p_t(\x_t)$, involving the time derivative $\frac{\diff p_t(x_t)}{\diff t}$. Integrating this equation, we will then express $p_1(\x_1)$ in terms of an initial density and a standard integral over time. We give this derivation in \cref{sec: likelihood}. It uses the fundamental and central Fokker--Planck equation, which governs the \emph{partial} time derivative of the marginal distribution: $\frac{\partial p_t(x_t)}{\partial t}$. It is discussed in \cref{sec: Fokker--Planck} and derived in \cref{sec: Kramers--Moyal}. %

\section{The Fokker--Planck equation}\label{sec: Fokker--Planck}

The \emph{Fokker--Planck equation} (FPE) is a partial differential equation (PDE) that describes the temporal evolution of the probability density $p_t(\x)$ of a state $\x$ in a stochastic or deterministic dynamical system.
A key feature of the FPE is that it provides a unified description of how these densities evolve, regardless of the specific ODE or SDE. Notably, different types of dynamics---SDEs, ODEs, and reverse-time SDEs---can all give rise to the same set of marginal distributions $p_t(\x)$, $t\in[0,1]$.
This equivalence underlies several modern generative modeling techniques. For example, diffusion models are typically formulated as learning to reverse an SDE that adds noise to data. Similarly, turning an SDE into an ODE and vice versa without changing the marginals $p_t(\x)$ enables deterministic sampling from a diffusion model and stochastic sampling from a deterministic flow model.

In this section, we introduce the Fokker--Planck equations for the differential equations in \cref{sec: ODE and SDE} and use it to define different processes with the same marginal densities $p_t$.
The FPE for the SDE \eqref{eq:SDE} is: 
\begin{align}
    \frac{\partial p_t(\x) }{\partial t} 
    &= - \sum\nolimits_i \frac{\partial}{\partial x_i} \left[ f_{i}(\x,t)  p_t(\x) \right] + \frac{1}{2} \sum\nolimits_{i,j} \frac{\partial^2}{\partial x_i \partial x_j}  \left[ \left[ \sigmab(\x,t) \sigmab\transp (\x,t)\right]_{ij} p_t(\x) \right] \\
    &= - \nabla \cdot \left[ \f(\x, t) p_t(\x) \right] + \frac{1}{2} \nabla \cdot \left( \nabla \cdot \left[ \D(\x, t) p_t(\x) \right] \right) \ ,
    \label{eq:Fokker Planck SDE}
\end{align}
where in the last line we defined the positive semi-definite diffusion matrix $\D(\x,t) \coloneqq \sigmab(\x,t) \sigmab(\x,t)\transp$ and used standard notation for divergences of vector and matrix fields.\footnote{Vector and matrix fields simply means vectors and matrices that are functions of $\x$. The divergence of a vector field $\v$ is defined as $\nabla \cdot \v \coloneqq \sum_i \frac{\partial v_i}{\partial x_i}$.  For a matrix field $\A$, we denote by $\nabla \cdot \A$ the vector field whose $i$-th element is the divergence of the $i$-th row of $\A$, i.e., $(\nabla \cdot \A)_i \coloneqq \nabla \cdot \A_i = \sum_j \frac{\partial}{\partial x_j}A_{ij}$.}
Notice that the drift $\f(\x, t)$ of the SDE corresponds to a first-order derivative, and the diffusion $\sigmab(\x,t)$ to a second-order derivative.
For a deterministic ODE, the FPE reduces to the \emph{Liouville} or \emph{continuity} equation, as the diffusion term vanishes:
\begin{align}
    \label{eq:Fokker Planck ODE}
    \frac{\partial p_t(\x)}{\partial t} 
    &= - \nabla \cdot \left[ \f(\x, t) p_t(\x) \right]  \ .
\end{align}
In the ODE setting, the drift is often referred to as the flow field, as it determines how probability mass is transported through state space.  

We derive the Fokker--Planck equation using the \emph{Kramers--Moyal expansion} in \cref{sec: Kramers--Moyal}. %
The derivation starts from the definition of the partial derivative and writes it as a Taylor expansion in the (jump) moments of the \emph{transition kernel} $p_{t+\Delta t}(\x',\x)$. For the Wiener process, it turns out that only the two first jump moments will contribute, leading to the drift and diffusion coefficient terms in the Fokker--Planck equation. An alternative equivalent derivation based upon Itô's lemma is also possible \citep{Risken1996}. %

\subsection{ODEs and SDEs with the same time-marginals}
\label{sec: from ODE to SDE with same marginals}

Consider the ODE \eqref{eq:ODE} with initial condition $\x_0 \sim p_0$, and let $p_t$ denote the marginal density of $\x_t$, satisfying the continuity equation \eqref{eq:Fokker Planck ODE}.
Here we will show that we can rewrite \cref{eq:Fokker Planck ODE} in a form that is similar to the Fokker--Planck equation \eqref{eq:Fokker Planck SDE}, which corresponds to an SDE. That will allow us to define an SDE with an arbitrary diffusion coefficient that, starting from $p_0$, induces the same time-marginals $p_t$ as the ODE.

\paragraph{From ODE to SDE.}
We can rewrite the diffusion term in the Fokker--Planck equation \eqref{eq:Fokker Planck SDE} as a drift term:
\begin{align}
    \nabla \cdot \left( \nabla \cdot \left[ \D(\x, t) p_t(\x) \right] \right) 
    &= \nabla \cdot \Bigg[ p_t(\x) \underbrace{\frac{1}{p_t(\x)} \nabla \cdot \left[ \D(\x,t) p_t(\x)\right]}_{\mathrm{(*)}} \Bigg]
    \label{eq: rewriting 2nd to 1st order FPE}
\end{align}
where $\mathrm{(*)}$ plays the role of a drift.
For any sufficiently regular $\sigmab$, we can add and subtract the same term to the r.h.s. of the continuity equation \eqref{eq:Fokker Planck ODE} to obtain a Fokker--Planck equation for an SDE:
\begin{align}
    \frac{\partial p_t(\x)}{\partial t} 
    &=  -\nabla \cdot \left[ \f(\x, t) \; p_t(\x) \right] \label{eq: FPE ODE} \\
    &=  -\nabla \cdot \Bigg[ \left(\f(\x,t) + \frac{1}{2 \, p_t(\x)}  \nabla \cdot \left[ \D(\x, t) p_t(\x) \right] \right) \; p_t(\x) \Bigg] + \frac{1}{2} \nabla \cdot \left( \nabla \cdot \left[ \D(\x, t) p_t(\x) \right] \right) \ .  \label{eq: FPE SDE sigma}
\end{align}
The two equivalent equations (\ref{eq: FPE ODE},\ref{eq: FPE SDE sigma}) %
correspond to the following ODE and SDE, respectively:
\begin{alignat}{3}
    \diff\x_t &= \f(\x_t,t) \,\diff t&&\qquad\qquad \text{(ODE)} \label{eq: ODE/SDE comparison sigma - ODE} \\
    \diff\x_t &= \left(\f(\x,t) + \frac{1}{2 \, p_t(\x)}  \nabla \cdot \left[ \D(\x, t) p_t(\x) \right] \right) \diff t+ \sigmab(\x_t,t) \, \diff\W_t &&\qquad\qquad \text{(SDE)}  \label{eq: same marginals ODE/SDE}
\end{alignat}
Since these two equations are derived from the same Fokker--Planck equation, they will have the same $p_t$ solution. 
The effect of the introduced diffusion $\sigmab$ is exactly canceled by the additional term in the drift.
So we can conclude that, \emph{given an initial distribution $\x_0 \sim p_0$ and an ODE \eqref{eq: ODE/SDE comparison sigma - ODE}, we can define an SDE \eqref{eq: same marginals ODE/SDE} with arbitrary diffusion coefficient $\sigmab(\x_t,t)$ that has the same time-marginals $p_t$ as the ODE}. In \cref{app: continuity preserves probability}, we further discuss the relationship between the ODE and SDE as continuity equations for probability. 

\paragraph{Reverse-time SDEs.} 

In diffusion models, we will use a diffusion process that is initialized to a data point $\x_1\approx\y$ at $t=1$ and runs backwards in time, gradually adding more noise. We will therefore need to extend the above analysis to the reverse-time case. 

We have just shown how to manipulate the Fokker--Planck equation to turn an ODE into an SDE with arbitrary stochasticity without changing the solution for the time-marginals $p_t(\x)$. The ODE can be solved forward and backward in time so we can use the results above to define a reverse-time SDE for the reverse-time solution to the ODE. Writing the reverse-time ODE in terms of the reverse clock $\tau$ that goes from 0 to 1 when $t$ goes from 1 to 0, $\tau \coloneqq 1-t$, we can introduce an SDE for clock $\tau$ that has the same marginals as the ODE. To formalize this, we write the original ODE $\frac{\diff\x_t}{\diff t}=\f(\x_t,t)$ in terms of clock $\tau$ renaming $\x_t$ to $\x'_\tau\coloneq \x_{1-\tau}=\x_t$, which yields $\frac{\diff\x'_\tau}{\diff \tau}=\f'(\x'_\tau,\tau)$. Since $\frac{\diff\x_t}{\diff \tau}=-\frac{\diff\x_t}{\diff t}$, the $\tau$ clock ODE has drift $\f'(\x, \tau) = -\f(\x,1-\tau)$. We can now trivially write down the reverse-time SDE for the clock $\tau$ that has the same marginals as the ODE:
\begin{alignat}{3}
\label{eq: RT ODE tau}
    \diff \x'_\tau & = \f'(\x'_\tau, \tau) \, \diff\tau \  && \text{(RT ODE, $\tau$ clock)} \\
    \label{eq: arbitrary reverse SDE (tau)}
    \diff\x'_\tau &= \left( \f'(\x'_\tau,\tau) + \frac{1}{2 \; p_t(\x'_\tau)} \nabla \cdot [ \D'(\x'_\tau, \tau) \, p_t(\x'_\tau) ]  \right)\, \diff\tau + \sigmab'(\x'_\tau,\tau) \, \diff\W'_\tau \hspace{2em}  && \text{(RT SDE, $\tau$ clock)}
\end{alignat}
with $\sigmab'(\x,\tau)=\sigmab(\x,1-\tau)$ and for simplicity we have kept $t$ in the marginal as a shorthand for $p_t(\x)=p_{1-\tau}(\x)$.

While this SDE is properly defined as running forward in the $\tau$ clock, it is common in the literature to re-express it in terms of the original time variable $t$, effectively describing the process backward in time. Here we provide a derivation sketch, while a detailed derivation is given in \cref{app: reverse-time SDE}.

The Wiener process $\W'_\tau$ introduced above starts from $\W'_0=\0$ and has independent increments according to the $\tau$ clock. This is a standard Wiener process running backward in time according to the $t$ clock, and can be defined in terms of the standard (forward) Wiener process as $\W'_\tau \coloneqq \W_{1-\tau} - \W_1$. We now change the variables in the reverse-time SDE above back to the $t$ clock and write 
\begin{align}
    \diff\x_t &= \left( \f(\x_t,t) - \frac{1}{2 \; p_t(\x_t)} \nabla \cdot \left[ \D(\x, t) \; p_t(\x_t) \right] \right) \diff t+ \sigmab(\x_t,t) \, \overleftarrow{\diff}\W_t \qquad\qquad\hspace{1em} \text{(RT SDE, $t$ clock)}  \label{eq: same marginals ODE/RT-SDE}
\end{align}
where we used the substitutions $\x'_\tau \to \x_t$, $\diff\tau\to-\diff t$, $\f'(\x'_\tau, \tau) \to -\f(\x_t,t)$ and $\sigmab'(\x'_\tau, \tau) \to \sigmab(\x_t,t)$. The left arrow on the stochastic differential denotes a so-called backward Itô integral; see \cref{app: reverse-time SDE} for further details.

We have now written the reverse-time SDE with marginals $p_t(\x)$ in terms of the standard $t$ clock. Its structure mirrors the standard SDE with the reverse-time Wiener process replacing the standard Wiener process and a changed sign in the drift correcting the diffusion term. That is, when running the SDE in reverse we need to reverse the contribution to the drift countering the diffusion. 

An even stronger result, due to 
\citet{anderson1982reverse}, says that when the forward and backward SDEs have the same time marginals \emph{and} the same exact diffusion coefficient $\sigmab(\x,t)$, \emph{the path distributions}---not just the marginals---\emph{in both directions are also identical}.

\paragraph{Summary.}

We have shown that, for any $\sigmab : \R^d \times [0,1] \to \R^d$ such that the SDEs have solutions, \emph{the ODE~\eqref{eq: ODE/SDE comparison sigma - ODE}, SDE~\eqref{eq: same marginals ODE/SDE}, and reverse-time SDE~\eqref{eq: same marginals ODE/RT-SDE} have the same time-marginals $p_t(\x)$}. In the diffusion models literature, the ODE~\eqref{eq: ODE/SDE comparison sigma - ODE} is typically called the \emph{probability flow ODE} (PF ODE) for the SDE~\eqref{eq: same marginals ODE/SDE}. In the remainder of the paper when we discuss more than one of these three models, we will denote ODE drift by $\f(\x_t,t)$ and the forward and reverse time with arrows:
\begin{align}
    \overrightarrow{\f}(\x_t,t) & = \f(\x, t) + \frac{1}{2\,p_t(\x)}  \nabla \cdot \left[ \D(\x, t) p_t(\x) \right] \label{eq: forward SDE drift} \\
    \overleftarrow{\f}(\x_t,t) & = \f(\x, t) - \frac{1}{2\,p_t(\x)}  \nabla \cdot \left[ \D(\x, t) p_t(\x) \right] = \overrightarrow{\f}(\x_t,t) - \frac{1}{p_t(\x)}  \nabla \cdot \left[ \D(\x, t) p_t(\x) \right] \ .  \label{eq: reverse SDE drift} 
\end{align}
The divergence term is typically rewritten as follows:
\begin{align}
    \frac{1}{p_t(\x)}  \nabla \cdot \left[ \D(\x, t) p_t(\x) \right] 
    =\nabla \cdot \D(\x, t) + \D(\x, t) \; \nabla \log p_t(\x)  \label{eq: SDE drift expanded divergence} 
\end{align}
to highlight the \emph{Stein score} $\nabla \log p_t(\x)$, which plays a central role in the score matching framework discussed in \cref{sec: score and flow matching}.

In the literature, diffusion matrices are typically scaled identity matrices, i.e., $\sigmab(\x, t) = \sigma(\x,t) \,\I$. In addition, $\sigma(\x,t)$ is typically chosen to be constant in space and is denoted as $\sigma_t$.
With these two assumptions, the drifts of the SDE and reverse SDE can be simplified as follows:
\begin{align}
    \f(\x, t) \pm \frac{1}{2\,p_t(\x)}  \nabla \cdot \left[ \D(\x, t) p_t(\x) \right] 
    &= \f(\x, t) \pm \frac{1}{2} \left( \nabla \sigma^2(\x,t) + \sigma^2(\x,t) \nabla \log p_t(\x) \right) \\
    &= \f(\x, t) \pm \frac{1}{2} \sigma_t^2 \; \nabla \log p_t(\x) \qquad\quad \mathrm{for} \ \,  \sigma(\x,t)=\sigma_t  \ .
\end{align}

\section{Maximum likelihood training with ODEs and SDEs} \label{sec: likelihood}

In the machine learning approach, the ODE/SDE and the likelihood function introduced in \cref{sec: Generative models from differential equations} are written in terms of a parameter vector $\thetab$. We can use neural networks to parameterize the drift $\f_{\thetab}$ and diffusion coefficient $\sigmab_{\thetab}$ that govern the latent process $\{\x_t\}_{t\in[0,1]}$ in $\R^d$ described by the ODE~\eqref{eq:ODE} or SDE~\eqref{eq:SDE}. The likelihood function $p_{\thetab}(\y|\x_1)$, mapping the terminal latent $\x_1$ to the observed variable $\y$, may also have parameter dependencies.

The objective is to perform \emph{maximum likelihood optimization} of the model using a training set $\{\y^{(m)}\}_{m=1}^n$. We want to use a gradient ascent method to maximize the log-likelihood $\sum_{m=1}^n \log p_{\thetab}(\y^{(m)})$, where $p_{\thetab}(\y) = \int p_{\thetab}(\y|\x_1)p_{1,\thetab}(\x_1)\diff\x_1$ is obtained from \cref{eq:likelihood} by adding the parameter dependencies.
This marginal likelihood $p_{\thetab}(\y)$ is typically intractable, i.e., it is not available in closed form and difficult to estimate using numerical methods. Furthermore, even if we could obtain samples from the posterior $p_{1,\thetab}(\x_1|\y)$, computing the log-likelihood gradient would still require that we can compute gradients of $\log p_{1,\thetab}(\x_1)$:
\begin{equation}
\nabla_{\thetab} \log p_{\thetab}(\y) = \mathbb{E}_{p_{1,\thetab}(\x_1|\y)} \big[ \nabla_{\thetab} \log p_{\thetab}(\y|\x_1) + \nabla_{\thetab} \log p_{1,\thetab}(\x_1) \big]  \ .
\end{equation} 
In the following, we will describe how to make gradient computations tractable. We omit the parameter dependence to keep the notation as simple as possible. Specifically, we introduce a variational posterior $q_1(\x_1|\y)$ (with parameters $\phib$) that approximates the true posterior $p_1(\x_1|\y) \propto p(\y|\x_1)p_1(\x_1)$. Note that the prior distribution for $\x_1$, $p_1(\x_1)$, typically will be a complicated distribution. 
In contrast, the posterior $p_1(\x_1|\y)$ is a much narrower distribution because conditioning on a data point will limit the support to only the part of latent space associated with the data point and its slight variations. We can therefore hope that choosing $q_1(\x_1|\y)$ as a simple distribution with support on the relevant part of latent space will be a good approximation to the true posterior. Furthermore, as discussed in the following, if $q_1(\x_1|\y)$ can be evaluated in closed form and efficiently sampled from, then we get a tractable lower bound on the log-likelihood. We discuss how to construct good variational distributions in \cref{sec: drift from flow} and the variational distribution used in diffusion, score and flow matching in \cref{sec: diffusion score flow}. 

A standard lower bound on the log-likelihood, the \emph{Evidence Lower BOund} (ELBO) is obtained from the following rewrite of the marginal log-likelihood:
\begin{align}
    \log p(\y)
    &= \E_{q_1(\x_1|\y)} \left[ \log p(\y) \right] \nonumber \\
    &= \E_{q_1(\x_1|\y)} \left[ \underbrace{\log \frac{p(\y | \x_1) p_1(\x_1)}{p_1(\x_1 | \y)}}_{\log p(\y)} + \underbrace{\log \frac{q_1(\x_1|\y)}{q_1(\x_1|\y)}}_{=0}\right] \nonumber \\
    &= \E_{q_1(\x_1|\y)} \left[ \log p(\y | \x_1)\right] - \E_{q_1(\x_1|\y)} \left[ \log \frac{q_1(\x_1|\y)}{p_1(\x_1)}\right] + \E_{q_1(\x_1|\y)} \left[ \log \frac{q_1(\x_1|\y)}{p(\x_1 | \y)}\right] \nonumber \\
    &= \underbrace{\E_{q_1(\x_1|\y)} \left[ \log p(\y | \x_1)\right] - \KL(q_1(\x_1|\y) \| p_1(\x_1))}_{\ODEELBO(\y)} + \underbrace{\KL(q_1(\x_1|\y) \| p_1(\x_1 | \y))}_{\text{ELBO gap ($\geq 0$)}} \ ,
    \label{eq: general ELBO with gap}
\end{align}
where $\KL(q(\x)\|p(\x)) \coloneqq \E_{q(\x)}\left[\log \frac{q(\x)}{p(\x)} \right]$ is the \emph{Kullback--Leibler (KL) divergence}, which is non-negative. We denote the ELBO as $\ODEELBO$ for reasons that will become clearer in \cref{sec: latent ODE}. 
The non-negative ELBO gap is a measure of the mismatch between the true and approximate posterior distributions: it is zero if and only if $q_1(\x_1|\y) = p_1(\x_1|\y)$, and in that case the ELBO coincides with the log-likelihood.

The first term in the ELBO is often referred to as the \emph{reconstruction term} since it involves ``encoding'' $\y$ into a latent variable $\x_1$ and then ``reconstructing'' $\y$ with a probabilistic model. Assuming that $q_1(\x_1|\y)$ is easy to sample from and that $\log p(\y|\x_1)$ can be evaluated, the reconstruction term can be easily estimated by Monte Carlo sampling.
For the second term, $\KL(q_1(\x_1|\y) \| p_1(\x_1)) = \E_{q_1(\x_1|\y)} \left[ \log q_1(\x_1|\y) - \log p_1(\x_1) \right]$, we can choose $q_1(\x_1|\y)$ such that it is easy to evaluate and sample from. However, $\log p_1(\x_1)$ is not available except in trivial cases. This is typically the main obstacle to estimating the ELBO, and the following sections develop two strategies that overcome this obstacle.

\textbf{Roadmap.}
In \cref{sec: latent ODE,sec: latent SDE}, we consider two approaches to address the issue of estimating the second term in the ELBO, $\KL(q_1(\x_1|\y) \| p_1(\x_1))$. This depends on how we define the latent process: (1) If the latent process is governed by an ODE, $\log p_1(\x_1)$ can be computed exactly by a change-of-variable formula, but this requires solving an ODE. Plugging this expression into the ELBO produces a \emph{simulation-based} training objective. (2) For a latent process described by an SDE, directly evaluating $\log p_1(\x_1)$ is not feasible in general. Instead, we will derive a bound on the second term in the ELBO that can be estimated without solving any differential equations, thus enabling efficient \emph{simulation-free} training.
Finally, in \cref{sec: mapping x1-y}, we discuss the role of the likelihood function $p(\y|\x_1)$ and the variational distribution $q_1(\x_1|\y)$, which define the link between the observable data $\y$ and the final-time latent variable $\x_1$.

\subsection{Latent ODE}
\label{sec: latent ODE}

If the latent process $\{\x_t\}_{t\in[0,1]}$ satisfies an ODE,
$\log p_1(\x_1)$ can be computed exactly, as we show next.
For the ODE $\frac{\dx}{\dt} = \f(\x, t)$, the \emph{total} time-derivative of the density (i.e., along a path) is:
\begin{align*}
     \frac{\diff}{\diff t} p_t(\x) 
     &= \frac{\partial p_t(\x)}{\partial t}   + \frac{\partial p_t(\x)}{\partial \x} \cdot \frac{\diff\x}{\diff t} \\
     &\!\eqnum{\ref{eq:Fokker Planck ODE}} - \nabla \cdot \left[ \f(\x,t) p_t(\x) \right] + \frac{\partial p_t(\x)}{\partial \x} \cdot \frac{\diff\x}{\diff t}  \\     
     & = - p_t(\x) \, \nabla \cdot \f(\x,t)  - \f(\x,t) \cdot \frac{\partial  p_t(\x)}{\partial \x}  + \frac{\partial p_t(\x)}{\partial \x} \cdot \frac{\diff\x}{\diff t}  \\
     & = - p_t(\x) \, \nabla \cdot \f(\x,t) + \frac{\partial  p_t(\x)}{\partial \x} \cdot \underbrace{\left( \frac{\diff\x}{\diff t} - \f(\x,t) \right)}_{=0 \ \mathrm{(ODE)}}
     = -p_t(\x) \, \nabla \cdot \f(\x,t) \ ,
\end{align*}
which can be rearranged into $\frac{\diff}{\diff t} \log p_t(\x) = - \nabla \cdot \f(\x,t)$. This equation can be combined with the state equation \eqref{eq:ODE} into an ODE in $d+1$ variables:
\begin{align}
    \label{eq: ODE log density}
     \frac{\diff}{\diff t} \begin{bmatrix}
    \x\\
    \log p_t(\x)
    \end{bmatrix}   & =   \begin{bmatrix}
    \f(\x,t)\\
    - \nabla \cdot \f(\x,t) 
    \end{bmatrix}
\end{align}
Let $\phib_s^t$ denote the \emph{flow map}, defined such that $\frac{\diff}{\diff t} \phib_s^t(\x_s) = \f(\phib_s^t(\x_s),t)$ and $\phib_t^t(\x)=\x$. Intuitively, this maps $\x_s$ to $\x_t$ along the ODE solution initialized at $\x_s$ at time $s$.
Integrating \cref{eq: ODE log density} on $[s,t]$, we obtain the following change of variable formula:
\begin{align}
    \log p_t(\phib^t_s(\x_s)) = \log p_s(\x_s) - \int_s^t \nabla \cdot \f(\phib_s^\tau(\x_s), \tau) \diff \tau \ ,
    \label{eq: ODE change of variable}
\end{align}
where $\nabla \cdot \f(\x', t)$ is shorthand for 
$\left(\nabla_{\x} \cdot \f\right)(\x', t)$, i.e., the spatial divergence evaluated at $\x'$.
In \cref{app: continuity preserves probability}, we give an interpretation of this result.

To compute $\log p_1(\x_1)$ for a given terminal value $\x_1$, we integrate the system of ODEs in \cref{eq: ODE log density} backwards from $t=1$ to $t=0$, obtaining $\x_0=\phib_1^0(\x_1)$ and $\log p_0(\x_0)$, and computing $\int_0^1 \nabla \cdot \f(\phib_1^t(\x_1), t) \, \diff t$ in the process. Substituting these quantities into \cref{eq: ODE change of variable} gives $\log p_1(\x_1)$. The ELBO \eqref{eq: general ELBO with gap} then takes the form:
\begin{align}
    \label{eq: ODE ELBO definition}
    \log p(\y)
    \geq \ODEELBO(\y)
    &\coloneqq \E_{q_1(\x_1|\y)} \left[ \log p(\y | \x_1)\right] - \KL(q_1(\x_1|\y) \| p_1(\x_1)) \\
    &= \E_{q_1(\x_1|\y)} \left[ \log p(\y | \x_1) - \log q_1(\x_1|\y) + \log p_0(\phib_1^0(\x_1)) - \int_0^1 \nabla \cdot \f(\phib_1^t(\x_1), t) \, \diff t\right]  \ . \nonumber
\end{align}
This objective serves as the basis for training. Because its evaluation requires solving the augmented ODE in \cref{eq: ODE log density}, it is often referred to as a \emph{simulation-based} objective in the generative modeling literature.
In certain cases, this objective coincides with the exact log-likelihood: for example, as we discuss in \cref{sec: mapping x1-y}, when the observation $\y$ is continuous and $\y=\x_1$, the bound becomes tight.
Generation also requires simulation, where samples are obtained by integrating the forward dynamics in \cref{eq:ODE} from prior to data \citep{grathwohl2018ffjord, chen2018neural}. At this stage, both training and inference depend on numerical ODE solvers, but in \cref{sec: latent SDE} we will discuss \emph{simulation-free} training.

To make the procedure concrete, we provide pseudo-code in \cref{alg:ODE ELBO}. For simplicity, we illustrate the algorithm using Euler discretization for integration, but higher-order methods such as Runge--Kutta are generally required for numerical stability, and gradients can be computed efficiently using the adjoint method \citep{chen2018neural,kidger2021hey}. For a broader discussion of numerical solvers in this setting, see \citet{kidger2021hey,karras2022elucidating}.

\begin{algorithm}%
\DontPrintSemicolon
  
\KwInput{$T$ (number of discretization steps), $\f$ (drift function), $p_0$ (prior), $p$ (likelihood function), $q$ (variational distribution), $\y$ (data)}

  \SetKwFunction{LogLike}{ODEELBO}
  \SetKwFunction{Sample}{Sample}
  \SetKwFunction{ODE}{ODESolver}
    
  \SetKwProg{Fn}{Def}{:}{}
  \Fn{\ODE{$\x_\mathrm{init}$, $t_\mathrm{init}$, $\Delta t$}}{ 
        $\x, t,  \mathrm{logP} = \x_\mathrm{init}, t_\mathrm{init}, 0$\;
         \For{i=1,\ldots,T}{        
            $\x = \x + \f(\x,t) \Delta t$\;
            $\mathrm{logP} = \mathrm{logP} - \nabla \cdot \f(\x,t) \Delta t$\;
            $t = t + \Delta t$
        }
        \KwRet $\x, \mathrm{logP}$\;
  }
  \SetKwProg{Fn}{Def}{:}{\KwRet}
  \Fn{\LogLike{$\y$}}{
        $\x_1 \sim q_1(\x_1|\y)$\;
        $\x_0, \mathrm{ELBO}$ = {\fontfamily{cmtt}\selectfont ODESolver}($\x_\mathrm{init}=\x_1$, $t_\mathrm{init}=1$, $\Delta t=-1/T$)\;
        $\mathrm{ELBO} = \mathrm{ELBO} + \log p_0(\x_0) + \log p(\y|\x_1) - \log q_1(\x_1|\y)$\;
        \KwRet $\mathrm{ELBO}$\;
  }

  \SetKwProg{Fn}{Def}{:}{\KwRet}
  \Fn{\Sample{}}{
        $\x_0 \sim p_0(\x_0)$\;
        $\x_1,\_$ = {\fontfamily{cmtt}\selectfont ODESolver}($\x_\mathrm{init}=\x_0$, $t_\mathrm{init}=0$, $\Delta t=1/T$)\;
        $\y_\mathrm{sample} \sim p(\y_\mathrm{sample}|\x_1)$\;
        \KwRet $\y_\mathrm{sample}$\;
  }
\caption{ODE pseudo-code for ELBO computation and sampling.}
\label{alg:ODE ELBO}
\end{algorithm}

\subsection{Latent SDE}
\label{sec: latent SDE}

If the latent process follows an SDE, $p_1(\x_1) = \int p(\x_1|\x_0) p(\x_0)\diff \x_0$ cannot be computed directly via a change of variables, and is therefore intractable in most cases. As a result, the KL divergence $\KL(q_1(\x_1|\y)\|p_1(\x_1))$ in the ELBO \eqref{eq: general ELBO with gap} is also intractable.
We will address this by deriving an upper bound to $\KL(q_1(\x_1|\y)\|p_1(\x_1))$ which consists in a time-averaged mean-squared error that can be efficiently estimated, as seen in diffusion models and flow matching (see \cref{sec: diffusion score flow}). This will result in a tractable, though generally looser, log-likelihood bound $\SDEELBO(\y) \leq \ODEELBO(\y) \leq \log p(\y)$. Consequently, unlike the ODE approach in \cref{sec: latent ODE}, the SDE formulation enables training without explicitly solving differential equations.

We assume the latent process is governed by the SDE
\begin{align}
    \dx = \overrightarrow{\f}(\x, t) \dt + \sigmab(\x, t) \diff\W 
\end{align}
and has marginals $p_t(\x)$. We define a variational latent process:
\begin{align}
    \dx = \overrightarrow{\g}(\x, \y, t) \dt + \sigmab(\x, t) \diff\W 
\end{align}
that is conditional on a data point $\y$ and has marginals $q_t(\x|\y)$.
We then rewrite the KL divergence as:
\begin{align}
    \KL(q_1(\x_1|\y) \,\|\, p_1(\x_1)) = \KL(q_0(\x_0|\y) \,\|\, p_0(\x_0)) + \int_0^1 \frac{\diff }{\diff t} \KL(q_t(\x|\y) \,\|\, p_t(\x)) \, \diff t \ .
\end{align}
The KL derivative can be written in terms of the partial derivatives of the two densities (see \cref{sec: derivative of kl divergence}):
\begin{align}
    \frac{\diff}{\dt} \KL(q_t(\x | \y) \,\|\, p_t(\x))
    &= \int \frac{\partial q_t(\x|\y)}{\partial t}\log \frac{q_t(\x|\y)}{p_t(\x)}\dx - \int \frac{q_t(\x|\y)}{p_t(\x)} \frac{\partial p_t(\x)}{\partial t} \dx \ .
\end{align}
Although $p_t(\x)$ and $q_t(\x|\y)$ come from SDEs, we will now work with their associated probability-flow ODE vector fields $\f$ and $\g$ since they satisfy simple continuity equations. These ODEs have the same time-marginals $p_t,q_t$ as the SDEs, and the relationship between their drifts is discussed in \cref{sec: from ODE to SDE with same marginals} (in particular, see \cref{eq: forward SDE drift}).
Plugging in these continuity equations, we obtain the following expression (proof in \cref{sec: derivative of kl divergence}):
\begin{align}
  \frac{\diff}{\diff t} \KL(q_t(\x | \y) \,\|\, p_t(\x))
    &= \E_{q_t(\x|\y)} \big[ (\g(\x, \y, t) - \f(\x, t)) \cdot (\nabla \log q_t(\x|\y) - \nabla \log p_t(\x)) \big] \ . 
    \label{eq: derivative of KL}
\end{align}
Using \cref{eq: reverse SDE drift},
we can also define a \emph{reverse-time} SDE with drift $\overleftarrow{\f}$ and same diffusion $\sigmab(\x,t)$ as the generative SDE above, that has the same time-marginals $p_t$. Similarly, we define $\overleftarrow{\g}$ from $\g$, using again the same diffusion $\sigmab$.
In \cref{sec: derivative of kl as diffusion losses}, we integrate \eqref{eq: derivative of KL} to show that %
\begin{align}
    \int_0^1 \frac{\diff}{\diff t}\KL\big(q_t(\x | \y) \,\|\, p_t(\x)\big) \,\dt
    &= \mathcal{D}(\overrightarrow{\f},\overrightarrow{\g},\y)   -  \mathcal{D}(\overleftarrow{\f},\overleftarrow{\g},\y) \ ,
\end{align}
where we assumed that $\D(\x,t)= \sigmab(\x,t)\sigmab\transp(\x,t)$ is invertible, and we defined
\begin{align}
    \mathcal{D}(\f,\g,\y) 
    &\coloneqq 
    \frac{1}{2} \int_0^1  \E_{q_t(\x\mid\y)}\left[ \left(\f(\x,t)- \g(\x,\y,t)\right)\transp \, \D(\x,t)^{-1} \, \left(\f(\x,t)- \g(\x,\y,t)\right) \right] \diff t \\
    &=\frac{1}{2}  \int_0^1 \E_{q_t(\x\mid\y)}\left[ \frac{\left\|\f(\x,t)- \g(\x,\y,t)\right\|^2}{\sigma^2(\x,t)}  \right] \diff t \ .
\end{align}
The last step follows from assuming an isotropic diffusion coefficient, i.e., $\D(\x, t) = \sigma^2(\x, t) \,\I$, and it results in the usual MSE loss we find in diffusion models and flow matching.

Using the decomposition derived above, the KL divergence at $t=1$ admits the exact representation:
\begin{align}
  \KL(q_1(\x_1|\y)\|p_1(\x_1)) = \KL(q_0(\x_0|\y)\|p_0(\x_0))  + \mathcal{D}(\overrightarrow{\f},\overrightarrow{\g},\y)   -  \mathcal{D}(\overleftarrow{\f},\overleftarrow{\g},\y) \ .
\end{align}
Substituting this into \eqref{eq: general ELBO with gap} yields a full decomposition of the marginal log-likelihood:
\begin{align}
    \log p(\y)
    &= \E_{q_1(\x_1|\y)} \left[ \log p(\y | \x_1)\right] -\KL(q_0(\x_0|\y)\|p_0(\x_0))  - \mathcal{D}(\overrightarrow{\f},\overrightarrow{\g},\y)   \label{eq: full elbo decomposition} \\
    & \qquad +  \mathcal{D}(\overleftarrow{\f},\overleftarrow{\g},\y) + \KL(q_1(\x_1|\y) \| p_1(\x_1 | \y)) \nonumber \ .
\end{align}
The first four terms on the r.h.s. correspond to the ELBO associated with the probability-flow ODE formulation (the ODE ELBO), while the final term is the usual variational gap (see \cref{eq: general ELBO with gap}).

The reverse-time discrepancy $\mathcal{D}(\overleftarrow{\f},\overleftarrow{\g},\y)$ is generally not available in closed form, since the reverse-time drifts depend on the (typically unknown) score of the generative model, $\nabla \log p_t$. However, $\mathcal{D}(\overleftarrow{\f},\overleftarrow{\g},\y)$ is always nonnegative because the quadratic form is weighted by $\D(\x,t)^{-1}$, which is positive definite.
Consequently, discarding the reverse discrepancy yields a valid (possibly looser) lower bound that remains tractable.
We therefore define the SDE-based ELBO as
\begin{align}
    \label{eq:ELBO_SDE}
    \SDEELBO(\y) &\coloneqq \underbrace{\E_{q_1(\x_1|\y)}\left[ \log p(\y|\x_1) \right]}_{\text{Reconstruction}} - \underbrace{ \KL(q_0(\x_0|\y) \| p_0(\x_0))}_{\text{Prior}}  - \underbrace{\mathcal{D}(\overrightarrow{\f},\overrightarrow{\g},\y)}_{\text{Diffusion}} \ .
\end{align}
With this notation, the decomposition \eqref{eq: full elbo decomposition} can be written compactly as
\begin{align}
    \log p(\y) &= \underbrace{\SDEELBO(\y) +  \mathcal{D}(\overleftarrow{\f},\overleftarrow{\g},\y)}_{\ODEELBO(\y)} + \KL(q_1(\x_1|\y) \| p_1(\x_1 | \y))  \ ,
\end{align}
which makes explicit that $\SDEELBO(\y) \leq \ODEELBO(\y) \leq \log p(\y)$.

\paragraph{Interpretation of the $\mathcal{D}$ functionals as path-space KL divergences.}
Above, we decomposed the KL divergence $\KL\big(q_1(\x | \y) \,\|\, p_1(\x)\big)$ into three tractable terms using only the continuity/Fokker--Planck equations and algebraic manipulations.
An equivalent, more classical derivation starts from standard identities for KL divergences between path measures (that is, the laws on continuous trajectories induced by the generative and variational SDEs). This yields two path-space KL terms, corresponding to forward and reverse processes. Evaluating these terms with Girsanov’s theorem gives precisely the functionals $\mathcal{D}(\overrightarrow{\f},\overrightarrow{\g},\y)$ and $\mathcal{D}(\overleftarrow{\f},\overleftarrow{\g},\y)$ obtained above, and hence exactly the decomposition in \cref{eq: full elbo decomposition}. This enriches the interpretation of our objective: the terms obtained by integrating the KL derivative are in fact information-theoretic quantities measuring discrepancies between trajectory laws. We present the more direct derivation in the main text because it uses only the Fokker--Planck equation and requires less background. In \cref{app: latent SDE path}, we give the standard path-space derivation in detail, including an informal derivation of Girsanov's theorem.

\paragraph{Simulation-free ELBO estimation.} An important property of the diffusion term is that it requires an expectation over $q_t(\x|\y)$. For a general choice of variational drift function $\g(\x,\y,t)$, samples from the time-marginal $q_t(\x|\y)$ can only be obtained by sampling an entire path from $q(\X|\y)$ which requires \emph{simulating} an SDE. However, as discussed below and in \cref{sec: drift from flow}, we can choose $q_t(\x|\y)$ to be a specific distribution (e.g., a Gaussian, or the transformation of a simple distribution through a normalizing flow) and derive the drift function corresponding to that choice of marginal. With this direct access to marginal samples, we can estimate the diffusion term by Monte Carlo sampling, drawing $t$ from the uniform distribution over $[0,1]$, denoted $\mathcal{U}[0,1]$, since $\int_0^1 \ldots\diff t= \E_{t \in \mathcal{U}[0,1]} \left[ \ldots \right]$. This means that we now have a \emph{simulation-free} objective to learn the model. Generation, on the other hand, of course still requires simulation of the SDE e.g. with the Euler--Maruyama discretization. \cref{alg:SDE ELBO} presents the pseudo-code.

\begin{algorithm}%
\DontPrintSemicolon
  
\KwInput{$T$ (number of discretization steps), $\f$ (drift function), $\g$ (variational drift function), $\sigmab$ (diffusion function), $p_0$ (prior), $p$ (likelihood function), $q$ (variational distribution), $\y$ (data)}

  \SetKwFunction{LogLike}{ODEELBO}
  \SetKwFunction{Sample}{Sample}
    \SetKwFunction{SDE}{SDESample}
  \SetKwFunction{LogLike}{SDEELBO}
  \SetKwFunction{Sample}{Sample}

  \SetKwProg{Fn}{Def}{:}{}
  \Fn{\SDE{$\x_\mathrm{init}$, $t_\mathrm{init}$, $\Delta t$}}{ 
        $\x, t = \x_\mathrm{init}, t_\mathrm{init}$\;
         \For{i=1,\ldots,T}{        
            $\Delta \W \sim \mathcal{N}(0,\Delta t\I)$\;
            $\x = \x + \f(\x,t) \Delta t + \sigmab(\x,t) \Delta \W$\;
            $t = t + \Delta t$
        }
        \KwRet $\x$\;
  }

  \SetKwProg{Fn}{Def}{:}{\KwRet}
  \Fn{\LogLike{$\y$}}{
        $t \sim \mathcal{U}[0,1]$\;
        $\x_0,\x_t,\x_1 \sim q_0(\x_0|\y),q_t(\x_t|\y),q_1(\x_1|\y)$\;
        $\mathcal{D}_t =\frac{1}{2}\left(\f(\x_t,t)- \g(\x_t,\y,t)\right)\transp \D(\x_t,t)^{-1} \left(\f(\x_t,t)- \g(\x_t,\y,t)\right)$\;
        $\mathrm{ELBO} = \log p(\y|\x_1) + \log p_0(\x_0) - \log q_0(\x_0|\y) - \mathcal{D}_t$\;
        \KwRet $\mathrm{ELBO}$\;
  }

  \SetKwProg{Fn}{Def}{:}{\KwRet}
  \Fn{\Sample{}}{
        $\x_0 \sim p_0(\x_0)$\;
        $\x_1$ = {\fontfamily{cmtt}\selectfont SDESample}($\x_\mathrm{init}=\x_0$, $t_\mathrm{init}=0$, $\Delta t=1/T$)\;
        $\y_\mathrm{sample} \sim p(\y_\mathrm{sample}|\x_1)$\;
        \KwRet $\y_\mathrm{sample}$\;
  }
\caption{SDE pseudo-code %
with Euler-Maruyama discretization in {\fontfamily{cmtt}\selectfont SDESample}.}
\label{alg:SDE ELBO}
\end{algorithm}

\subsection{Mapping between latents and observables}
\label{sec: mapping x1-y}

In \cref{sec: latent ODE,sec: latent SDE}, we have established two approaches for estimating the ELBO---evaluating $\log p_1(\x_1)$ exactly via the ODE change-of-variables, or bounding the marginal KL by a tractable path KL, which results in a looser bound. In this section, we turn to a complementary issue: how the latent process interfaces with data through the likelihood function $p(\y | \x_1)$ and the variational posterior $q_1(\x_1 | \y)$.
While the design of these distributions is in principle orthogonal to whether the latent dynamics follow an ODE or an SDE, there are subtle dependencies. In the ODE case, we require $q_1(\x_1 | \y)$ to be easy to evaluate and sample from. In the SDE case, this requirement extends to all intermediate marginals $q_t(\x_t|\y)$, and these marginals must be compatible with the variational SDE \eqref{eq: variational SDE} with drift $\g(\x,\y,t)$.

The variational posterior $q_1(\x_1|\y)$ should be chosen with two properties in mind: tractability of sampling and density evaluation, and, ideally, closeness to the true posterior. If one were able to set $q_1(\x_1|\y) = p_1(\x_1|\y) \propto p_1(\y|\x_1)p_1(\x_1)$, then the ELBO gap in \cref{eq: general ELBO with gap} would vanish and the ELBO would coincide with the log-likelihood.\footnote{In the SDE approach, the corresponding statement holds at the path level: if $q(\X|\y)=p(\X|\y)$, then the SDE ELBO coincides with the ODE ELBO.}
In the deterministic case, $p(\y|\x_1) = \delta(\y-\x_1)$, this condition is automatically satisfied: the true posterior $p_1(\x_1|\y) \propto \delta(\y-\x_1) p_1(\x_1)$ is concentrated at $\x_1=\y$, and the variational posterior can be defined as the same Dirac distribution: $q_1(\x_1|\y) \coloneqq p_1(\x_1|\y) = \delta(\y - \x_1)$.\footnote{To show this more rigorously, we integrate the true posterior against a suitable test function $\phi$ and apply Bayes' theorem:
\begingroup
    \setlength{\abovedisplayskip}{4pt}%
    \setlength{\belowdisplayskip}{4pt}%
    \[
        \int \phi(\x_1) p_1(\x_1|\y) \diff\x_1
        = \frac{\int \phi(\x_1) p(\y|\x_1) p_1(\x_1) \diff\x_1}{\int p(\y|\x'_1) p_1(\x'_1) \diff\x'_1}
        = \frac{\phi(\y) p_1(\y)}{p_1(\y)} = \phi(\y)
    \]%
\endgroup
which holds for all $\phi$ and therefore implies $p_1(\x_1|\y) = \delta(\y - \x_1)$.
}
For latent ODE models this implies that the marginal log-likelihood $\log p(\y)$ is in fact computable exactly. This construction underlies continuous normalizing flows \citep{chen2018neural}, where the ODE dynamics transform a simple prior at $t=0$ into the observed data at $t=1$. In the SDE setting, the situation is more delicate: even if the final-time posterior is matched, there typically remains a gap due to mismatch between the true and variational posterior path distributions ($q(\X|\y) \neq p(\X|\y)$).

\section{Parameterization of the SDE model}\label{sec: drift from flow}

In this section, we discuss how to parameterize the SDE model in a way that ensures maximum flexibility while at the same time keeping the objective simulation-free, that is, being able to draw samples from the variational marginal distribution $q_t(\x|\y)$ at any time $t$ without simulating the variational SDE. We explore marginals defined as normalizing flows and mixtures of normalizing flows and derive their corresponding ODEs and SDEs. We discuss likelihood functions for continuous and ordinal data and how the variational distribution can be chosen to match the likelihood function. In \cref{sec: diffusion score flow}, we connect the results of this section to the diffusion literature. We present a numerical example in \cref{sec: experiments}.

\paragraph{Generative drift and the diffusion coefficient.} The SDE generative model is defined by the drift $\overrightarrow{\f}(\x,t)$ and the diffusion coefficient $\sigmab(\x,t)$, both of which can be directly parameterized by neural networks.

\paragraph{Variational drift.} 
In the following, we will work with ODE and forward-/reverse-time SDE variational drifts and denote these by $\g(\x,\y,t)$, $\overrightarrow{\g}(\x,\y,t)$ and $\overleftarrow{\g}(\x,\y,t)$, respectively. The definition of the variational drift $\overrightarrow{\g}(\x,\y,t)$ requires more care because we need to define it such that the marginal distribution $q_t(\x|\y)$ (that is the solution to the associated Fokker--Planck equation) admits computationally efficient sampling so that we get the benefit from 
the simulation-free ELBO as outlined in \cref{alg:SDE ELBO}. We will describe two routes to achieving this: start with the desired parameterization of the marginal and then derive the corresponding drift (this section), or use a Gaussian Markov chain that has Gaussian probability paths  \citep{ho2020denoising,kingma2021variational,nielsen2023diffenc,song2020score} (\cref{sec: diffusion score flow}). Below we will consider the marginal $q_t(\x_t|\y)$ being defined as a normalizing flow and a mixture of normalizing flows, and derive the corresponding explicit expression for the variational ODE drift $\g(\x,\y,t)$. Once we have expressions for both the marginal and the ODE drift, we can use the result in \cref{eq: same marginals ODE/SDE} to write the SDE drift as
\begin{equation}
    \label{eq: variational drift same marginals}
    \overrightarrow{\g}(\x,\y,t) = \g(\x,\y,t) + \frac{1}{2 \, q_t(\x|\y)}  \nabla \cdot \left[ \D(\x, t) q_t(\x|\y) \right] \ .
\end{equation}

\paragraph{Drift from marginal defined as normalizing flow.} We consider the marginal distribution $p_t(\x_t)$ induced by a flow $\F_t : \R^d \times [0,1] \to \R^d$ that is continuously differentiable in both variables \citep{bartosh2024neural}. Specifically, we will use this to construct the variational $q_t(\x_t|\y)$ from a flow $\G_t(\x;\y)$.

Given a random variable $\epsilonb$ with density $p(\epsilonb)$, we let $\x_t \coloneqq \F_t(\epsilonb)$ and assume $\F_t$ is invertible for all $t$, such that $\x_t = \F_t(\F_t^{-1}(\x_t))$. 
Here, $\epsilonb$ can be thought of as a \emph{Lagrangian label} in fluid dynamics, i.e., a unique identifier assigned to a particle to track its movement over time. 
The explicit expression for the marginal is given by the change-of-variable formula:
\begin{equation}
        p_t(\x) = \left| \det \frac{\partial \F^{-1}_t(\x) }{\partial \x} \right| p\left( \F^{-1}_t(\x) \right) \ . \label{eq: drift from flow: change of var}
\end{equation}
The derivation will be valid for general flows and base measures $p(\epsilonb)$. A concrete example that is relevant for practical applications is a linear flow with independent components: $\x_t = \F_t(\epsilonb) = \alphab(t) + \beta(t) \epsilonb$, with $p(\epsilonb)$ a standard Gaussian. The inverse is $\epsilonb = \F^{-1}_t(\x)= (\x-\alphab(t))/\beta(t)$. Below we generalize the derivation to a mixture of normalizing flows: $p_t(\x_t) = \sum_k \pi_k p_{k,t}(\x_t)$ with fixed mixture coefficients $\pi_k$. 

Now we define the velocity given by the flow $\F_t$:
\begin{align}
    \label{eq:drift from flow}
    \f(\x, t) \coloneqq \frac{\diff}{\diff t} \F_t(\epsilonb)  \quad\text{with}\quad \epsilonb = \F_t^{-1}(\x) \ .
\end{align}
Then, for each label $\epsilonb$, using the definition $\x_t \coloneqq \F_t(\epsilonb)$ we have
$ \frac{\diff}{\diff t} \x_t = \frac{\diff}{\diff t} \F_t(\epsilonb) = \f(\x, t) $,
i.e., $\x_t$ solves the ODE with velocity $\f$ and initial condition $\x_0 = \F_0(\epsilonb)$. This shows that, given a base density $p(\epsilonb)$ and a flow $\F_t$ that transports it to the desired marginals $p_t$ (\cref{eq: drift from flow: change of var}), we can derive the velocity of the associated ODE with \cref{eq:drift from flow}.
For the linear transformation: $\F_t(\epsilonb)= \alphab(t) + \beta(t) \epsilonb$, the ODE drift becomes $\f(\x_t,t) = \frac{\partial \alphab(t)}{\partial t} + \frac{\partial \beta(t)}{\partial t} \frac{\x_t-\alphab(t)}{\beta(t)}$.

\paragraph{Drift from marginal defined as a mixture.} For the mixture model $p_t(\x) = \sum_k \pi_k p_{k,t}(\x)$, each marginal $p_{k,t}(\x)$ is defined by a normalizing flow $\F_{k,t}(\x)$ with corresponding drift $\f_k(\x,t)$ obtained from \eqref{eq:drift from flow}: $\f_k(\x,t) =   \left. \frac{\diff \F_{k,t}(\epsilonb)}{\diff t} \right|_{\epsilonb=\F^{-1}_{k,t}(\x)}$. To derive the drift for the mixture we use the continuity equation for the mixture and for each component:
\begin{equation}
    \frac{\partial}{\partial t} p_t(\x) = \sum\nolimits_k \pi_k \frac{\partial}{\partial t} 
 p_{k,t}(\x_t) = - \sum\nolimits_k \pi_k \nabla \cdot \left[ \f_k(\x,t) p_{k,t}(\x) \right] = -\nabla \cdot \bigg[ \underbrace{  \sum\nolimits_k \pi_k \f_k(\x,t) p_{k,t}(\x) }_{ \f(\x,t) \, p_t(\x)}  \bigg] \ .
\end{equation}
This leads to
\begin{equation}
    \label{eq: ODE drift mixture}
    \f(\x,t) =  \sum\nolimits_k r_k(\x,t) \f_k(\x,t)  
\end{equation}
with the ``responsibility'' for component $k$ given input $\x$:
\begin{equation}
    r_k(\x,t) = \frac{\pi_k p_{k,t}(\x_t)}{\sum_{k'} \pi_{k'} p_{k',t}(\x_t)} \ . 
\end{equation}
This result can also be generalized to time-dependent mixture coefficients and the SDE setting \citep{brigo2002general,peluchetti2021non,skreta2025the}.

\paragraph{Variational drift example.} We will now use the methodology developed above to the standard choice of the variational distribution, a Gaussian: $q_t(\x_t|\y)=\mathcal{N}(\x_t|\alphab(\y,t),\beta^2(y,t)\I)$. The ODE drift becomes $\g(\x_t,\y,t) = \frac{\partial \alphab(\y,t)}{\partial t} + \frac{\partial \beta(\y,t)}{\partial t} \frac{\x_t-\alphab(\y,t)}{\beta(\y,t)}$ and the score $\nabla \log q_t(\x_t|\y) = - \frac{\x_t-\alphab(\y,t)}{\beta^2(\y,t)}$. From \eqref{eq: variational drift same marginals}, setting $\D(\x,t)=\sigma^2_t\I$, we get
\begin{equation}\label{eq: g linear}
    \overrightarrow{\g}(\x,\y,t) = \g(\x,\y,t) + \frac{\sigma^2_t}{2} \nabla \log q_t(\x|\y) = \frac{\partial \alphab(\y,t)}{\partial t} + \left( \frac{\partial \beta(\y,t)}{\partial t} - \frac{\sigma^2_t}{2\beta(\y,t)} \right) \frac{\x-\alphab(\y,t)}{\beta(\y,t)} \ . 
\end{equation}
We will investigate this drift function for a simple numerical example in Section \ref{sec: experiments}.
The extension to the mixture with marginals $q_{k,t}(\x_t|\y)=\mathcal{N}(\x_t|\alphab_k(\y,t),\beta_k(\y,t)\I)$ leads to the ODE drift \eqref{eq: ODE drift mixture} and a simple responsibility-weighted score function: $\nabla \log q_t(\x|\y) = \sum_k r_k(\x,\y,t)\nabla \log q_{k,t}(\x|\y) $ and thus the same form for the SDE drift: $\overrightarrow{\g}(\x,\y,t) =  \sum_k r_k(\x,\y,t) \overrightarrow{\g}_k(\x,\y,t)$.

\paragraph{Likelihood function examples.} We will consider likelihood functions for continuous and ordinal data. For continuous data, it is common to use a Gaussian with a simple covariance: 
\begin{equation}
    p(\y|\x) = \mathcal{N}(\y|\x,\delta^2\I) \ .
\end{equation}
For ordinal data, e.g., 8-bit pixels $\y \in[0,1,\ldots,255]^d$, we can take independent dimensions and build a discrete normalized distribution by the following construction \citep{kingma2021variational}: 
\begin{equation}
    p(\y|\x) = \prod_{i=1}^d  \frac{\mathcal{N}(y_i|x_i,\delta^2)}{ \sum_{y'} \mathcal{N}(y'|x_i,\delta^2)} \ . 
\end{equation}
With the choice $q_t(\x_t|\y)=\mathcal{N}(\x_t|\alphab(\y,t),\beta^2(t)\I)$, we can easily calculate the prior and reconstruction terms in the SDE ELBO \eqref{eq:ELBO_SDE}. 

We will often match the variational distribution at $t=1$ with the likelihood function, setting $\alphab(\y,1)=\y$ and $\beta(1)=\delta$ \citep{kingma2021variational}, where $\delta$ is set to a small fixed value or is a learned parameter. 
We will consider an explicit example in the context of flow matching in \cref{sec: flow matching}.

\section{Diffusion, score and flow matching}\label{sec: diffusion score flow}

Diffusion \citep{sohl2015deep,ho2020denoising,kingma2021variational}, score \citep{vincent2011connection,song2020score} and flow matching \citep{lipman2022flow,liu2022flow} are a set of closely related and extremely popular latent process generative models that can be well-understood in terms of the variational framework described in this paper. One may even say that your favorite score and flow matching objectives are secretly ELBOs. In this section, we will introduce each method and show how each method's learning objective can be identified as a special case of the SDE ELBO \eqref{eq:ELBO_SDE}.

\subsection{Diffusion models}\label{sec: variational diffusion}

The denoising diffusion probabilistic model (DDPM, \citet{ho2020denoising}) derives its name from being defined in terms of two processes: the diffusion, a reverse-time process initialized in the data that gradually adds noise so that at $t=0$ most of the information about the data is removed, and the denoising, a forward-time process that starting from pure noise gradually transforms the latent such that at $t=1$ one ideally gets samples from the training set distribution. The diffusion and denoising processes are readily identified with the variational and generative processes described above.     
The original diffusion formulation is defined in terms of a finite number of stochastic layers, that is, $\Delta t$ is finite, but it has a well-defined SDE limit \citep{kingma2021variational,song2021maximum}. We will give a brief derivation of this using the variational diffusion models (VDM) framework \citep{kingma2021variational}, which is mathematically equivalent to the DDPM.

\paragraph{Variational diffusion model (VDM)} In the VDM, the variational marginal is selected to be Gaussian:
\begin{equation}
q_t(\x_t|\y) =\mathcal{N}(\x_t|a_t\y,b_t\I) \ .
\end{equation}
that is a special case of the more general Gaussian setting  from the previous section with $\alphab(\y,t)=a_t\y$ and $\beta(\y,t) =b_t$.
Furthermore, a finite step size \emph{reverse-time} Markov process $q(\x_s|\x_t)$, $s<t$ is \emph{selected} that preserves the form of the marginal, that is, the marginal $q_s(\x_s|\y) = \int q(\x_s|\x_t) q_t(\x_t|\y) d\x_t$ is given by $\mathcal{N}(\x_s|a_s\y,b^2_s\I)$. The explicit form of the Markov process is
\begin{equation}
    q(\x_s|\x_t)=\mathcal{N}(\x_s|a_{s|t}\x_t, b_{s|t}^2\I) \ ,  \qquad a_{s|t}=\frac{a_s}{a_t} \quad \mathrm{and} \quad b_{s|t}^2= b^2_s-a^2_{s|t} b_t^2 \ . 
\end{equation}
This reverse-time Markov process is only defined when the signal-to-noise ratio $\SNR(t) \coloneqq \frac{a^2_t}{b^2_t}$ is a strictly increasing function with $t$. This can be achieved with a learned (monotonic) function or a fixed function such as linear in $\log\SNR(t)= t\, \log\SNR(1) + (1-t)\, \log\SNR(0)$. As a further restriction, a so-called variance-preserving process is often used, such that $a^2_t + b^2_t = 1$. \citet{song2020score} discusses various alternatives in the same class. We discuss the choice of denoising process after taking the SDE limit of the variational process.

\paragraph{SDE limit.} We can convert the finite step process to a difference equation and then take the $\Delta t \to 0$ limit like we did in the ELBO derivation in \cref{app: girsanov}. We write $q(\x_s|\x_t)$ as a reverse-time finite step size process:
\begin{equation}
    \Delta \x_t = \x_s -\x_t = a_{s|t} \x_t  + b_{s|t} \epsilonb - \x_t = \underbrace{\frac{a_{s|t}-1}{\Delta t} \x_t}_{\overleftarrow{\g}(\x,t)} \Delta t + \underbrace{\frac{b_{s|t}}{\sqrt{|\Delta t|}}}_{\sigma_t} \overleftarrow{\Delta}\W'_t
\end{equation} with $\Delta t = s-t$, $\epsilonb\sim \mathcal{N}(\epsilonb|\0,\I)$ and $\overleftarrow{\Delta}\W'_t$ being the finite step reverse-time Wiener process with variance $\Delta t$. We can derive the drift and diffusion coefficient in the limit:
\begin{align}
    \overleftarrow{\g}(\x,t) & =\frac{a_{s|t}-1}{\Delta t}\x_t=\frac{a_s-a_t}{\Delta t} \frac{\x_t}{a_t }\to \frac{\partial \log a_t}{\partial t}\x_t\\
    \sigma_t^2 & = \frac{b^2_{s|t}}{|\Delta t|} = \frac{\SNR(t)-\SNR(s)}{|\Delta t|}\frac{b_s^2}{\SNR(t)} \to b_t^2 \frac{\partial \log \SNR(t)}{\partial t} \ . 
\end{align}
We can use the result from Section \ref{sec: from ODE to SDE with same marginals}: $\overrightarrow{\g}(\x,\y,t) = \overleftarrow{\g}(\x,\y,t) + \sigma_t^2 \nabla \log q_t(\x|\y)$  to derive the forward SDE drift 
\begin{equation}
\overrightarrow{\g}(\x,\y,t)= \frac{\partial \log a_t}{\partial t}\x_t + \sigma_t^2 \nabla \log q_t(\x|\y) = \frac{\partial \log a_t}{\partial t}\x_t - \sigma_t^2 \frac{\x-a_t \y}{b^2_t} \ .
\end{equation}
This can be compared to the more general result using the linear normalizing flow \eqref{eq: g linear}.

\paragraph{Parameterizing generative drift as a signal prediction network.} Finally, it is preferred to parameterize the generative drift function $\overrightarrow{\f}(\x,t)$ in terms of a signal or noise prediction neural network. For example, one may define the drift as $\overrightarrow{\f}(\x,t)\coloneq \overrightarrow{\g}(\x,\widehat{\y}(\x,t),t)$ with $\widehat{\y}(\x,t)$ being the signal prediction neural network and $\overrightarrow{\g}(\x,\y,t)$ being the variational drift function. For the Gaussian variational used in the VDM this leads to a simplified expressions for the diffusion term in the SDE ELBO \eqref{eq:ELBO_SDE} with a signal from latent objective:
\begin{align}
    \mathcal{D}(\overrightarrow{\f},\overrightarrow{\g},\y) =  & \frac{1}{2}
     \int_0^1 \frac{\partial \SNR(t)}{\partial t} \E_{q_t(\x|\y)}\left[ || \y - \widehat{\y}(\x,t)||^2 \right]\diff t \\  =  & \frac{1}{2}
     \int_{\SNR(0)}^{\SNR(1)} \left. \E_{q_{t}(\x|\y)}\left[ || \y - \widehat{\y}(\x,t||^2 \right] \right|_{t=t(\SNR)}\diff \SNR \ , \nonumber
\end{align}
where in the second equality we have made a change of integration variable to $\SNR$ and $t(\SNR)$ is the inverse function of $\SNR(t)$. 

\subsection{Score matching}\label{sec: score and flow matching}

Score matching \citep{vincent2011connection,song2020score} derives its name from the objective of fitting $\s(\x,t)$, an approximation to the unconditional score function $\nabla \log p_t(\x)$, to the conditional score function $\nabla \log q_t(\x|\y)$. In this section, we will show that the score matching objective can be obtained from the diffusion term in the SDE ELBO for a specific choice of parameterization of the generative and variational processes. The reconstruction and prior terms in the ELBO can be ignored as these only depend upon the variational process which is fixed in the score matching framework. The score matching framework is thus optimizing the ELBO with respect to the generative process for a given variational distribution. This relationship between the SDE ELBO and the score matching has been pointed out in the literature before, e.g. by \citet{song2021maximum}. The original derivation of score matching takes a different route based upon an infinite dataset argument. It is presented in \cref{app:score and flow}.

Score matching uses the same reverse-time variational SDE as used in the diffusion model and again we can use \cref{sec: from ODE to SDE with same marginals} to reverse the direction of time:
\begin{equation}
    \label{eq: g score}
    \overrightarrow{\g}(\x,\y,t) = \overleftarrow{\g}(\x,t) + \nabla \sigma^2(\x,t) + \sigma^2(\x,t)\nabla \log q_t(\x|\y) \ .
\end{equation}
Likewise, we can write the generative drift as: $\overrightarrow{\f}(\x,t) = \overleftarrow{\f}(\x,t) + \nabla \sigma^2(\x,t) + \sigma^2(\x,t) \nabla \log p_t(\x)$. In order to get to a tractable score matching objective, we select $\overleftarrow{\f}(\x,t) = \overleftarrow{\g}(\x,t)$ and replace the unknown score function with a learnable score function $\s(\x,t)$:
\begin{equation}
    \label{eq: f score}
    \overrightarrow{\f}_{\mathrm{score}}(\x,t) = \overleftarrow{\g}(\x,t) + \nabla \sigma^2(\x,t) + \sigma^2(\x,t) \s(\x,t) \ .
\end{equation}
Since the score function only approximates the true drift, the transformation between the different drifts are only approximate as well. 
Inserting the score matching parameterization for the generative and variational SDE drifts equations \eqref{eq: f score} and \eqref{eq: g score} into the SDE ELBO diffusion term leads to the score matching objective:
\begin{align}
    \mathcal{D}_{\mathrm{Score}} 
    &= \frac{1}{2} \int_0^1 \E_{q_t(\x|\y)}\left[ \sigma^2(\x,t) \left\| \s(\x, t)- \nabla \log q_t(\x| \y)\right\|^2 \right]\diff t\ . \label{eq: diff loss as denoising score matching}
\end{align}
This is exactly the score-matching loss used in the diffusion models literature \citep{song2020score}, which is a weighted average of denoising score matching objectives \citep{vincent2011connection} over different noise levels. In score matching, the variational distribution is chosen in the same way as in diffusion models.

It can be shown, see Appendix \ref{app:score and flow}, that in the infinite-data limit the optimal $\s(\x,t)$ is the score function of the unconditional marginal distribution: $\s^*(x,t) = \nabla \log q_t(\x)$, where $q_t(\x)=\mathbb{E}_{q(\y)}[q_t(\x|\y)]$. Since the data distribution $q(\y)$ is not available, it is replaced by an average over the training set in the objective. The name score matching thus is derived from the objective of matching the data conditional score function $\nabla \log q_t(\x|\y)$ to a function that approximates the unconditional score function $\s(\x,t) \approx \nabla \log q_t(\x)$. 

\subsection{Flow matching} \label{sec: flow matching}

In flow matching \citep{lipman2022flow,liu2022flow} we use the same setup as in score matching, but instead of introducing the score as our objective we will use the flow (or velocity) which is another name for the ODE drift $\g(\x,\y,t)$. From Section \ref{sec: from ODE to SDE with same marginals}, we can formally write $\g(\x,\y,t) = \left( \overrightarrow{\g}(\x,\y,t) + \overleftarrow{\g}(\x,\y,t) \right) /2$. If we restrict the reverse-time drift to be data-independent we can rewrite the relationship between the drift terms as:
\begin{equation}
    \label{eq: g flow}
     \overrightarrow{\g}(\x,\y,t) = 2 \g(\x,\y,t) - \overleftarrow{\g}(\x,t) \ .
\end{equation}
Similarly, we may choose to define the generative SDE drift in terms of a flow function $\v(x,t)$:
\begin{equation}
    \label{eq: f flow}
     \overrightarrow{\f}_{\mathrm{flow}}(\x,t)  = 2 \v (\x,t) - \overleftarrow{\g}(\x,t)\ .
\end{equation}
Inserting these parameterizations into the diffusion term leads to the flow matching objective:
\begin{align}
    \mathcal{D}_{\mathrm{Flow}} 
    &= 2\int_0^1  \E_{q_t(\x|\y)}\left[\frac{\left\| \v(\x,t) - \g(\x,\y,t) \right\|^2}{\sigma^2(\x,t)} \right]\diff t \ . 
    \label{eq: diff loss as denoising flow matching}
\end{align}
In flow matching, samples of $\x_t$ are often obtained as $\x_t = t\y + (1-t)\epsilonb$ with $\epsilonb \sim \mathcal{N}(\epsilonb|\0,\I)$ or equivalently $q_t(\x|\y) = \mathcal{N}(\x|t\y,(1-t)^2\I)$. This type of interpolation can be generalized to other settings, see for example \citet{albergo2023stochastic,tong2023simulation}. We can get further insight about the flow matching objective, by using the result from \cref{sec: drift from flow} to write the variational drift in a particular simple form:
\begin{equation}
    \g(\x_t,\y,t) = \frac{\y-\x_t}{1-t} = \frac{y-t \y - (1-t)\epsilonb}{1-t} = \y -\epsilonb \ .  
\end{equation}
We can conclude from this that the flow matching objective's target for the drift $\v(\x_t,t)= \v(t\y + (1-t)\epsilonb,t)$ is $\y-\epsilonb$ with $\epsilonb \sim \mathcal{N}(\x_0|\0,\I)$. 

The diffusion loss derived above is all we need as an objective because the prior and reconstruction terms have no tunable parameters when we use the stochastic interpolation variational. Nonetheless, it is still instructive to write the prior and reconstruction terms to get the full ELBO picture. We will generalize the interpolation slightly to $x_t = ty + \beta_t \epsilonb$ with $\beta_t =(1-t)\beta_0+t\beta_1$. The prior term is $\KL(q_0(\x_0|\y) \| p_0(\x_0))= -\frac{d}{2} (\log \beta_0^2 + 1-\beta_0^2)$ for a standard Gaussian prior. The reconstruction term is $\E_{q_1(\x_1|\y)}\left[ \log p(\y|\x_1) \right]=-\frac{d}{2} (\log 2 \pi \delta^2 + \frac{\beta_1^2}{\delta^2})$ for the Gaussian likelihood $\mathcal{N}(\y|\x_1,\delta^2\I)$. We observe that reconstruction term diverges in the $\delta\to 0$ (noise-free) limit. The ELBO is thus ill-defined in that case. Optimizing $\beta_0$ and $\beta_1$ might lead to a higher ELBO than the flow matching defaults \citep{zheng2023improved}.  Optimizing the flow matching objective with respect to the diffusion coefficient $\sigma(\x,t)$ will lead to the trivial $\sigma(\x,t)\to \infty$ solution. This is directly connected to the ELBO gap diverging in this limit (results not shown).  

As shown in Appendix \ref{app:score and flow}, analogous to the score matching result, in the infinite data limit, the optimal flow corresponds to a marginalized drift:
$\v^*(\x,t) =  \mathbb{E}_{q_t(\y|\x)}[\g(\x,\y,t) ]$ 
with $q_t(\y|\x) = q_t(\x|\y) q(\y) / q_t(\x)$.
Flow matching may, like score matching be used either as a generative ODE using the learned flow function $\f(\x,t)=\v(\x,t)$ or as an SDE with drift $\overrightarrow{\f}(\x,t) = 2 \v(\x,t) - \overleftarrow{\g}(\x,t)$.

To sum up, score and flow matching may be viewed as slightly restricted versions of the general variational SDE approach because the variational is defined as a ``from data to noise'' diffusion (reverse-time SDE). The score and flow matching approaches have the advantage that learned generative model can be used both in an ODE and SDE formulation. This result is approximate because it hinges on a derivation of the objective which is only strictly valid in the infinite-data limit.

\section{Experiments} \label{sec: experiments}

We consider a toy example of learning a one-dimensional density defined as a mixture of $k=5$ Laplace distributions that has locations at $-2(k-1),\ldots,2(k-1)$ and scale equals to one for all components. We can draw data from the distribution to get a precise estimate of the expected log-likelihood $\int p(y) \log p(y) dy$. This gives $-2.9862$ for a sample of $n_{\mathrm{val}}=\num{8000}$ validation point. This is the reference value to compare with for the different trained models. All models are trained on the same data set of $n=\num{1000}$ data points. Jupyter notebooks to reproduce the results are available at \href{https://github.com/olewinther/generative-ode-sde}{github.com/olewinther/generative-ode-sde}

\begin{figure}
\begin{center}
\includegraphics[width=0.49\textwidth]{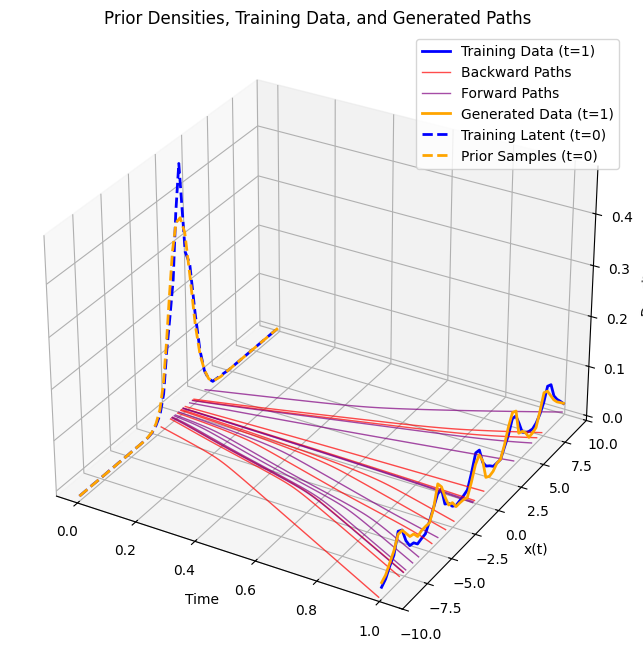}
\includegraphics[width=0.49\textwidth]{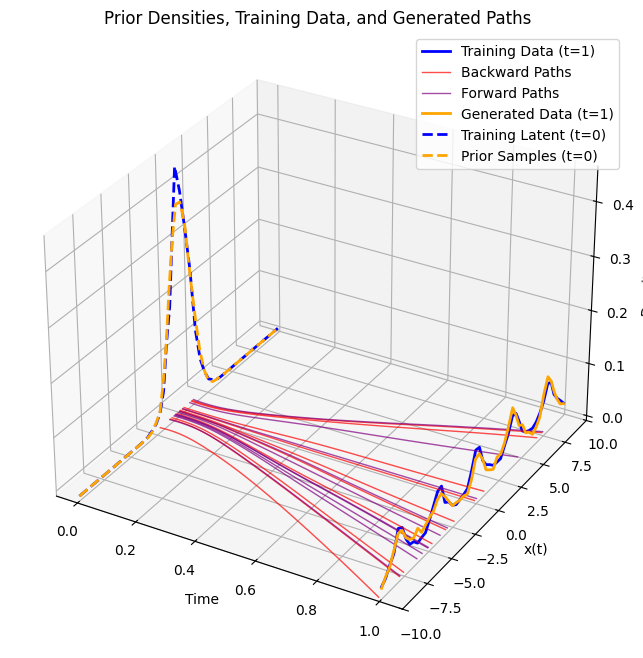}
\end{center}
\caption{Latent ODE (left) and flow matching (right). The ODE training results has log-likelihoods -3.025 (validation) and -2.990 (training). The blue densities are for the training data and yellow generated from the trained model. The red paths are for solving the ODE for some data point backwards in time and purple are generated data paths.  
}
\label{fig:ODE}
\end{figure}

We consider the ODE with a Dirac delta likelihood function. The prior distribution is set to $p_0(x) = \mathcal{N}(x|0,1)$. The ODE drift $f(x,t)$ is chosen to be a 2-64-64-1 feed-forward neural network with tanh activation functions for the hidden units and linear output unit. The Adam optimizer is run for 200 epochs with a learning rate of $10^{-2}$ and batch size of 125. The ODE uses $T=100$ discretization steps and a second-order midpoint Runge-Kutta integrator. The results for the ODE are shown in Figure \ref{fig:ODE} left. We run standard flow matching (\cref{sec: flow matching}) with $\sigma(x,t)=1$ (equal weighting of all times), learning rate of $10^{-3}$, batch size of 125 and increase the number of epochs to $\num{10000}$ due to cheaper and higher variance stochastic evaluation. The results for the flow matching are shown in Figure \ref{fig:ODE} right. As can be visually inspected, the two methods produce very similar solutions. Evaluating the ELBO for flow matching is possible, but will not give a value that can be compared to the other methods because we have not optimized the parameters of the likelihood and diffusion coefficient is not identifiable for the flow matching objective. 

\begin{figure}
\begin{center}
\includegraphics[width=0.49\textwidth]{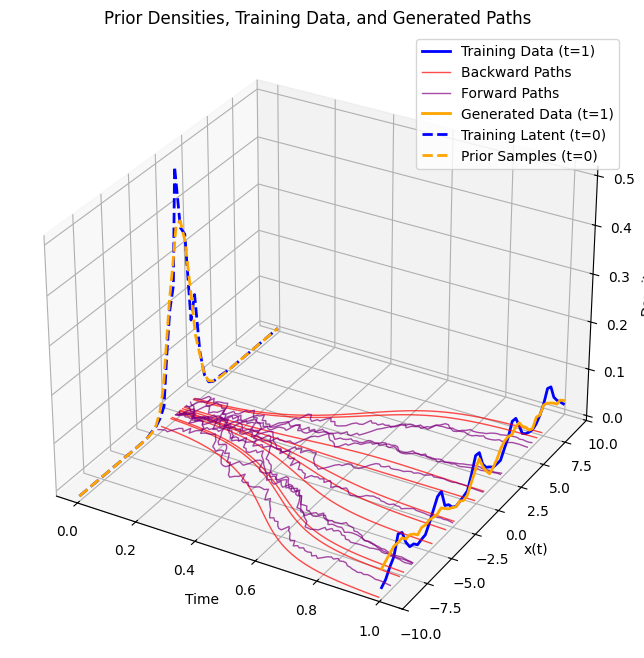}
\includegraphics[width=0.49\textwidth]{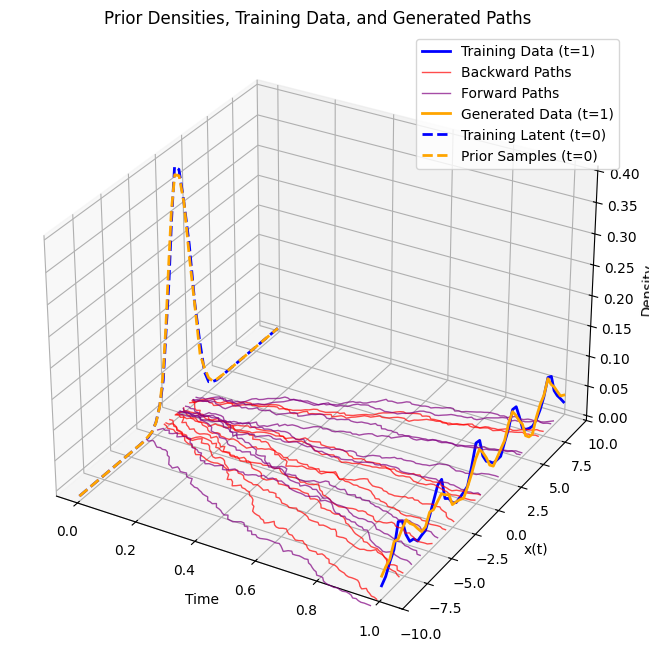}\\
\includegraphics[width=0.49\textwidth]{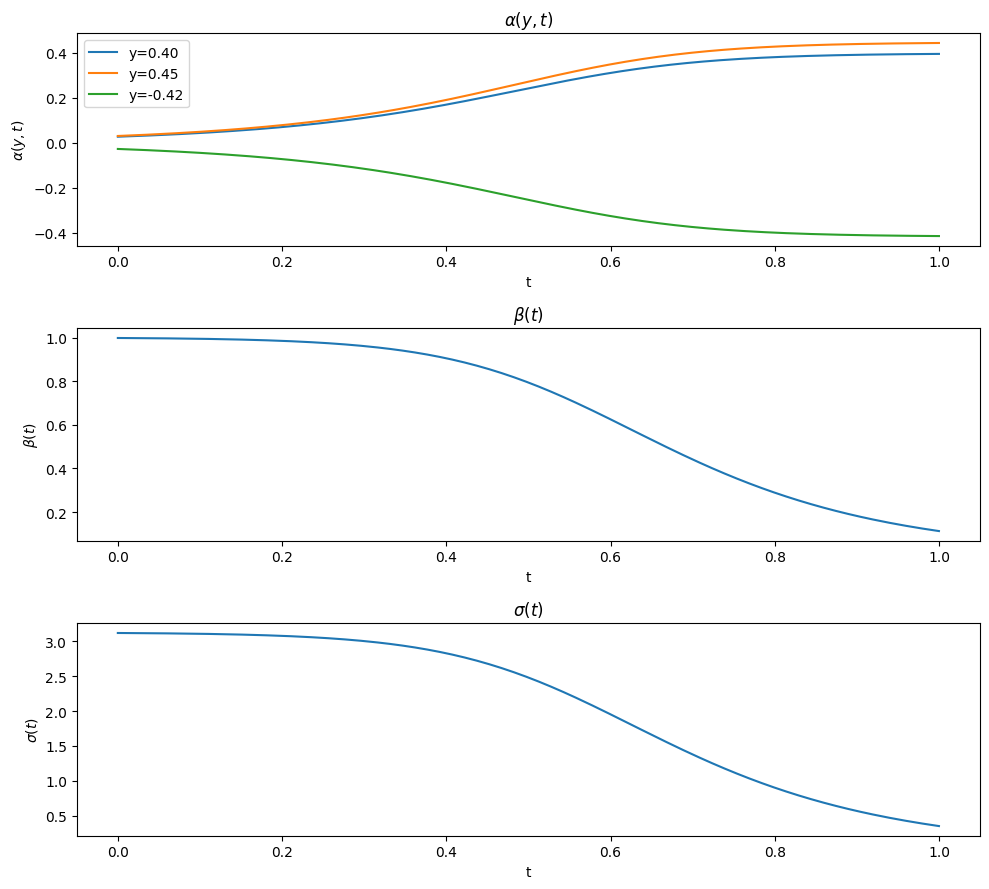}
\includegraphics[width=0.49\textwidth]{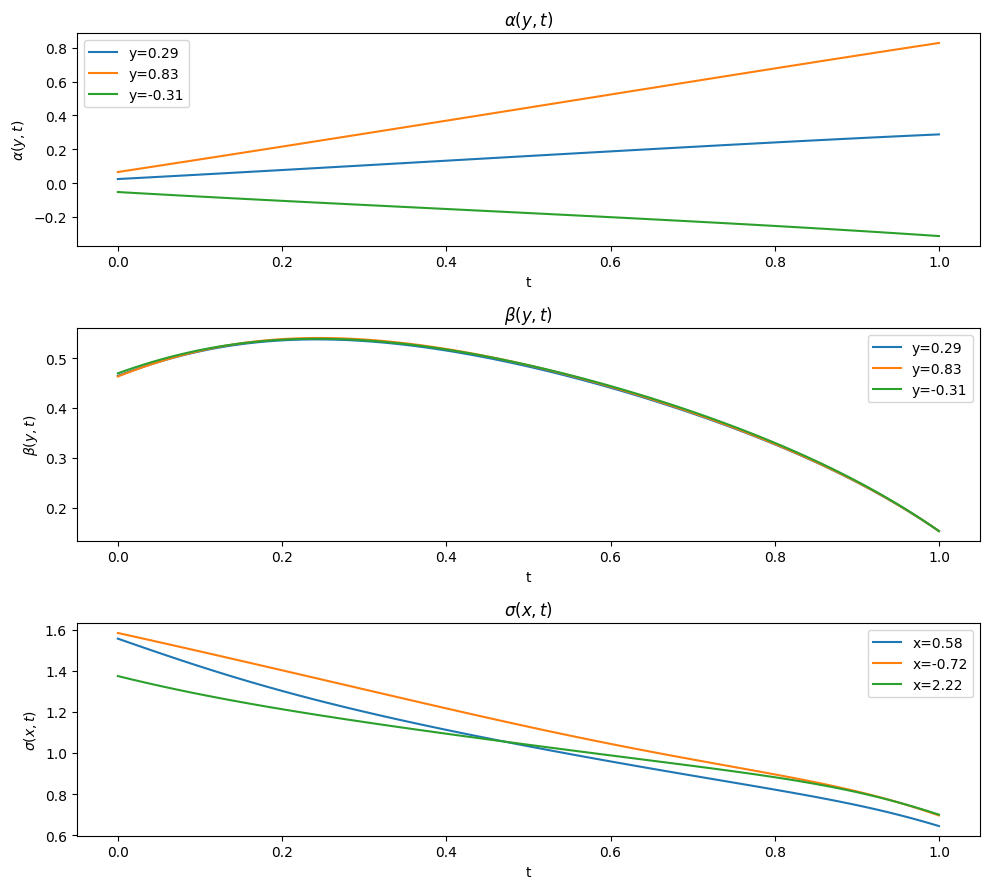}
\end{center}
\caption{SDE result with variational diffusion model (VDM) left and general parameterisation (right). For VDM, the smoothed ELBO values are -3.256 (validation) and -3.259 (training) and for the general parameterisation -3.052 (validation) and -2.975 (training). The lower plots show the learned functions. For the VDM, the backward varitional ODE paths are shown.  
}
\label{fig:SDE}
\end{figure}

For the SDE we also use the Gaussian prior. The likelihood function is $p(y|x_1)=\mathcal{N}(y|x_1,\delta^2)$ with $\delta$ being a non-negative learned parameter. The generative drift for the SDE $f(x,t)$ is the same architecture as used for the ODE. The diffusion network $\sigma(x,t)$ is a 2-64-64-1 feed-forward neural network with tanh activation functions for the hidden units and softplus on the output unit. The variational is chosen to be Gaussian: $q_t(x|y)=\mathcal{N}(\alpha(y,t),\beta^2(y,t))$. We consider two different parameterizations of the variational distribution. The first is the variational diffusion model (VDM, \cref{sec: variational diffusion}), with a linear scale for the log SNR and learned end-points. The likelihood and $t=1$ variational widths are matched: $\delta=b_1$. The signal prediction network is a 2-64-64-1 feed-forward network with tanh activation functions and linear output. In the second parameterisation we learn the variational freely apart from matching at $t=1$: $\alpha(y,t) = y t + (1-t) \widehat{\alpha}(y,t)$ and $\beta(y,t) = \delta t + (1-t) \widehat{\beta}(y,t)$ with $\widehat{\alpha}(y,t)$ being a 2-64-64-1 feed-forward neural network with tanh activation functions for the hidden units and linear output unit and $\widehat{\beta}(y,t)$ being a 2-64-64-1 feed-forward neural network with tanh activation functions for the hidden units and softplus output unit.The Adam optimizer is run for $\num{2000}$ epochs with a learning rate of $10^{-3}$ and batch size of 125. For sampling from the trained model $T=100$ discretizations steps were used. The results for these two setups are shown in Figure \ref{fig:SDE}. We observe that both approaches learn similar densities, but with the VDM converging slower with higher variance, with a lower validation ELBO and quite different learned $\alpha(y,t)$, $\beta(y,t)$ and $\sigma(x,t)$.

\section{Discussion} \label{sec: discussion}

This paper has provided a self-contained and informal methodological introduction to generative models based on ordinary and stochastic differential equations. The differential equations and the Fokker--Planck equation governing the temporal evolution of the marginal distribution of the latent variables have been presented and derived. The variational approach has been used to derive lower bounds on the log likelihood for the different models. This ELBO (evidence lower bound) serves as a tractable objective function for learning the parameters of the generative differential equation specified in terms of the drift and diffusion coefficient. Recently proposed popular machine learning methods for generative modeling with differential equations (neural ordinary differential equations, diffusion, score and flow matching models) have been presented in the context of the general variational approach. 

The purpose of this paper is to present the relevant methodology in one place. Various practical considerations necessary for practical use, such as neural architectures for efficient inference and good performance, will have to be found in the recent more application-focused literature.

The application of generative modeling in machine learning has reached a very mature stage in a relatively short time with very impressive results in diverse areas ranging from image and video to molecules. Interestingly, the setting used in machine learning is most often that of a simple (conditional) prior followed by a latent path and then generation. The typical setting in statistics, where the differential equation describes a latent process for a \emph{time series}, is rarely considered in the machine learning setting. This stems from the fact that construction of a suitable variational latent process is much more difficult than in the ``one observation'' case. Going beyond the one observation scenario is thus a largely unexplored research topic that could have implications for further use of variational techniques for modeling time series \citep{bartosh2025sde}.

\clearpage

\bibliography{tmlr}
\bibliographystyle{tmlr}

\clearpage

\appendix

\section{The Kramers--Moyal expansion and the Fokker--Planck equation} \label{sec: Kramers--Moyal}
 
In this appendix we show (i) a general Taylor series expansion expression for the partial time derivative of the marginal density that (ii) for the SDE will only consist of the first and second order term. %
The Fokker--Planck equation holds for any continuous-time stochastic process as long as there are no jumps, i.e., discrete changes of the stochastic variables \citep{Risken1996,oksendal2013stochastic,gardiner2009stochastic}. Markov jump processes \citep{gardiner2009stochastic} provide a tool to deal with jumps in the process, but this is beyond the scope of this paper. 

\subsection{Kramers--Moyal}

The Fokker--Planck equation is a special case of a more general equation, the Kramers--Moyal expansion, that describes the evolution of the density $p_t(\x)$ over time in any stochastic process. In this section, we will derive the Kramers--Moyal expansion, and in \cref{sec:sec_fp_sde} we will show that for Wiener (Brownian) stochastic processes it has only two terms and reduces to the Fokker--Planck equation. %

The derivation of the Kramers--Moyal expansion starts from the definition of the partial derivative:
\begin{equation}
\label{eq:def_pde}
    \frac{\partial p_t(\x) }{\partial t} = \lim_{\Delta t \to 0} \frac{1}{\Delta t} \left( p_{t+\Delta t}(\x) - p_t(\x)  \right) \ ,
\end{equation}
where we denote the finite time difference by $\Delta t$ and take its limit to zero. 
The time-shifted distribution $p_{t+\Delta t}(\x)$ can be expressed as a marginalization over the distribution of $\x$ at time $t$:
\begin{equation}
\label{eq:marg}
    p_{t+\Delta t}(\x) = \int p_{t+\Delta t}(\x|\x') p_t(\x') \diff\x' \ ,
\end{equation}
where we will leave the ``transition kernel'' $p_{t+\Delta t}(\x|\x')$ unspecified for now. In Section \ref{sec:sec_fp_sde} we consider stochastic differential equations that can be discretized to give a Gaussian transition kernel. 
Now, for any smooth analytic function $h(\x)$ such that $h(\x) \frac{\partial p_t(\x) }{\partial t}$ is integrable, we can use \cref{eq:def_pde,eq:marg} to get:
\begin{align}
    \int h(\x) \frac{\partial p_t(\x) }{\partial t} \diff\x 
    &\eqnum{\ref{eq:def_pde}} \int_\x h(\x) \left[ \lim_{\Delta t \to 0} \frac{1}{\Delta t} \left( p_{t+\Delta t}(\x) - p_t(\x)  \right) \right] \diff \x  \label{eq: integral dp/dt h}\\
    &\eqnum{\ref{eq:marg}} \lim_{\Delta t \to 0} \frac{1}{\Delta t} \left( \int_\x h(\x) \int_{\x'} p_{t+\Delta t}(\x|\x') p_t(\x') \diff\x' \diff\x  - \int_{\x'} h(\x') p_t(\x') \diff\x' \right) \nonumber\\
    &=  \lim_{\Delta t \to 0} \frac{1}{\Delta t} \bigg( \int_\x h(\x) \int_{\x'} p_{t+\Delta t}(\x|\x') p_t(\x') \diff\x' \diff\x  \nonumber\\
    &\qquad 
    - %
    \int_{\x'} h(\x') p_t(\x') \underbrace{\int_\x p_{t+\Delta t}(\x|\x') \diff\x}_{=1} \diff\x' \bigg)\nonumber\\
    &= \lim_{\Delta t \to 0} \frac{1}{\Delta t} \int_\x \int_{\x'} p_{t+\Delta t}(\x'|\x) p_t(\x) (h(\x') -h(\x) ) \diff\x'\diff\x \ , \nonumber
\end{align}
where in the last step we have swapped names of the integration variables $\x$ and $\x'$. 
Applying a Taylor expansion of $h(\x')$ around $\x$, 
we can rewrite \cref{eq: integral dp/dt h} as follows:
\begin{align}
    \int h(\x) \frac{\partial p_t(\x) }{\partial t} \diff\x 
    &= \lim_{\Delta t \to 0} \frac{1}{\Delta t}  \int_\x \int_{\x'} p_{t+\Delta t}(\x'|\x) p_t(\x) (h(\x') -h(\x) ) \diff\x'\diff\x \\
    &= \lim_{\Delta t \to 0} \frac{1}{\Delta t}  \int_\x \int_{\x'} p_{t+\Delta t}(\x'|\x) p_t(\x) %
    \sum_{n=1}^\infty \sum_{i_1,\ldots,i_n=1}^d \frac{(x'_{i_1}-x_{i_1})\cdots(x'_{i_n}-x_{i_n})}{n!} \nonumber \\ 
    & \mbox{\hspace{7.5cm}} \times \underbrace{\frac{\partial^n h(\x)}{\partial x_{i_1}\cdots\partial x_{i_n}}}_{=h^{(n)}_{i_1,\ldots,i_n}(\x)}
    \diff\x'\diff\x \nonumber\\
    &= \int p_t(\x) \sum_{n=1}^\infty \sum_{i_1,\ldots,i_n=1}^d D^{(n)}_{i_1,\ldots,i_n}(\x,t) \; h^{(n)}_{i_1,\ldots,i_n}(\x) \diff\x \ , \nonumber
\end{align}
where we defined the \textit{Kramers--Moyal coefficients} $D^{(n)}_{i_1,\ldots,i_n}(\x,t)$ for $n \in \N^+$ (also called jump moments) as:
\begin{align}
    D^{(n)}_{i_1,\ldots,i_n}(\x,t) 
    &\equiv \frac{1}{n!} \lim_{\Delta t \to 0} \frac{1}{\Delta t} \int (x'_{i_1}-x_{i_1})\cdots(x'_{i_n}-x_{i_n}) p_{t+\Delta t}(\x'|\x) \diff\x' \label{eq: jump moments definition} \\
    &= \frac{1}{n!} \lim_{\Delta t \to 0} \frac{1}{\Delta t} \E[\Delta x_{t,i_1} \cdots \Delta x_{t,i_n} | \x_t] \ , \label{eq: jump moments with expectation}
\end{align}
where the conditional expectation of the increment $\Delta \x_t=\x_{t+\Delta t}-\x_t$ is over the distribution $p_t(\Delta \x_t|\x_t)$.

With integration by parts, we shift the derivatives from $h(\x)$ to $p_t(\x) D^{(n)}_{i_1,\ldots,i_n}(\x,t)$. To show how this works consider the first order term:
\begin{align}
   \sum_{i=1}^d \int p_t(\x) D^{(1)}_i(\x,t) \frac{\partial h(\x)}{\partial x_i} \diff\x = \sum_{i=1}^d \underbrace{\left[ p_t(\x) D^{(1)}_i(\x,t) h(\x) \right]_{-\infty}^{+\infty}}_{=0\, \mathrm{by\, assumption}} - \int h(\x) \sum_{i=1}^d\frac{\partial}{\partial x_i} \left[  D^{(1)}_i(\x,t) p_t(\x)\right] \diff\x
\end{align}
The assumption of the first term being zero is based upon the following argument: the density $p_t(\x)$ must necessarily vanish at $\pm \infty$ because it integrates to 1 and the jump moment $D^{(1)}_i(\x,t)$ and $h(\x)$ (that we choose) are both well-behaved so that they together will not grow faster than $p_t(\x)$ goes to zero.
For the $n$-th order term we apply this $n$ times to arrive at:
\begin{equation}
    \int h(\x) \frac{\partial p_t(\x) }{\partial t} \diff\x = \int h(\x) \sum_{n=1}^\infty (-1)^n \sum_{i_1,\ldots,i_n} 
    \frac{\partial^n}{\partial x_{i_1}\cdots\partial x_{i_n}}
    \left[ D^{(n)}_{i_1,\ldots,i_n}(\x,t) p_t(\x) \right] \diff\x \ .
\end{equation}
Since this has to hold for any $h$, we get the Kramers--Moyal expansion:
\begin{equation}
    \label{eq:KramersMoyal}
    \frac{\partial p_t(\x) }{\partial t} = \sum_{n=1}^\infty (-1)^n \sum_{i_1,\ldots,i_n} 
    \frac{\partial^n}{\partial x_{i_1}\cdots\partial x_{i_n}}
    \left[ D^{(n)}_{i_1,\ldots,i_n}(\x,t) p_t(\x) \right] \ ,
\end{equation}
which is an infinite sum that either truncates at $n=2$ or has infinitely many nonzero terms \citep[Pawula's theorem;][]{pawula1967approximation,Risken1996}.

Comparing this equation with the Fokker--Planck equation (\ref{eq:Fokker Planck SDE}) we see that only including the two first orders gives the Fokker--Planck equation and we identify the drift coefficient $f_i(\x,t) = D^{(1)}_i(\x,t)$ and the diffusion coefficient $\sum_k\sigma_{ik}(\x,t)\sigma_{jk}(\x,t)/2 = D_{ij}^{(2)}(\x,t)$ with the two first drift moments. In the next section we will show how SDEs are connected to the Fokker--Planck equation.

\subsection{Derivation of Fokker--Planck for the SDE}
\label{sec:sec_fp_sde}

We can derive the Fokker--Planck equation for the SDE starting from the Kramers--Moyal expansion Equation (\ref{eq:KramersMoyal}) and show that only the two first jump moments will be non-zero in the infinitesimal limit. 

We can calculate the jump moments Equation (\ref{eq: jump moments with expectation}) for the SDE by noting that the conditional for the increment $p( \diff\x|\x_t,t)$ used in the definition of the jump moment is Gaussian for the SDE because we use the Wiener process:
\begin{equation}
    p_t(\diff\x \mid \x) =\mathcal{N}(\diff\x \mid \f(\x,t)\diff t, \;\sigmab(\x,t)\sigmab\transp (\x,t)\diff t) \ . 
\end{equation}
Using the definition can immediately write down the two first jump moments: $D^{(1)}_i(\x,t) = f_i(\x,t)$ and $D^{(2)}_{ij}(x,t) = \sum_k\sigma_{ik}(\x,t)\sigma_{jk}(\x,t)/2$. To show that the higher moments disappear we will calculate

$\E[d x_{t,i_1} \cdots d x_{t,i_n}  \mid  \x_t]$ which is just a non-central moment of a Gaussian with mean $m_i=f_i(x,t)\diff t$ and covariance $v_{ij} = \sum_k\sigma_{ik}(\x,t)\sigma_{jk}(\x,t) \diff t$. Since we are only interested in the scaling property in the infinitesimal limit, we can simplify to counting powers of mean $m$ and variance $v$ components. The non-central moment is a sum where in our case only one term dominates as $dt \to 0$:
\begin{alignat}{3}
    n \text{ odd}: &\qquad \E[\diff  x_{t,i_1} \cdots \diff  x_{t,i_n}  \mid  \x_t] &&= mv^{\frac{n-1}{2}} + o(mv^{\frac{n-1}{2}}) = O(\diff t \cdot (\diff t)^{\frac{n-1}{2}}) = O((\diff  t)^{\frac{n+1}{2}}) \\
    n \text{ even}: &\qquad \E[\diff  x_{t,i_1} \cdots \diff  x_{t,i_n}  \mid  \x_t] &&= v^{\frac{n}{2}} + o(v^{\frac{n}{2}}) = O((\diff t)^{\frac{n}{2}}) \nonumber
\end{alignat}
or in a single expression using the ceiling function:
\begin{align}
    \E[\diff  x_{t,i_1} \cdots \diff x_{t,i_n} | \x_t] &= O((\diff t)^{\lceil\frac{n}{2}\rceil}) \ .
\end{align}
Now when taking the limit in the definition of the jump moment we divide by $\diff  t$, so any term with $n > 2$ will vanish, since it will be $O((\diff t)^{\lceil\frac{n}{2}\rceil - 1})$ where the exponent is $\geq 1$. So the SDE recovers the Fokker--Planck Equation (\ref{eq:Fokker Planck SDE}) with $D^{(1)}_i(\x,t) = f_i(\x,t)$ and $D^{(2)}_{ij}(\x,t) = \sum_k\sigma_{ik}(\x,t)\sigma_{jk}(\x,t)/2$.

\section{Reverse-time SDEs}
\label{app: reverse-time SDE}

Here we give the derivation of the reverse-time SDE using the forward clock $t$ stated in \cref{sec: from ODE to SDE with same marginals}
While this can be made rigorous (see, e.g., \citet{kunita2019stochastic}),
we will present here a heuristic derivation by leveraging the view of SDEs as limits of discrete-time processes.

We start by introducing the \emph{reverse-time Wiener process} $\W'_t \coloneqq \W_{1-t} - \W_1$. Because this construction simply reverses the original trajectory and shifts it so that it starts at 0, this is a standard Wiener process when viewed backward in time.
We also need to define $\sigmab(\x, t) \coloneqq \sigmab'(\x, 1-t)$ and
\begin{align}
    \overleftarrow{\f}(\x, t) 
    &\coloneqq - \overrightarrow{\f'}(\x, 1-t) \\
    &= - \f'(\x,1-t) - \frac{1}{2 \; p'_{1-t}(\x)} \nabla \cdot \left[\D'(\x,1-t) \; p'_{1-t}(\x) \right] \\
    &= \f(\x,t) - \frac{1}{2 \; p_t(\x)} \nabla \cdot \left[ \D(\x, t) \; p_t(\x) \right] \ .
\end{align}
Now, note that the precise meaning of \cref{eq: arbitrary reverse SDE (tau)} is given by the integral equation:
\begin{align}
    \x'_\tau &= \x'_0 + \int_0^\tau \overrightarrow{\f'}(\x'_s,s) \, ds + \int_0^\tau \sigmab'(\x'_s,s) \, \diff\W_s \ .
\end{align}
Consider its discretization with steps $\{\nicefrac{i}{T}\}_{i=0}\transp$ and rewrite it in reverse-time by reindexing with $j = T-i$:
\begin{align}
    \x_{\frac{k}{T}} = \x'_{1-\frac{k}{T}}
    &= \x'_0 + \frac{1}{T} \sum_{i=0}^{T-k-1} \overrightarrow{\f'}(\x'_{\frac{i}{T}},\nicefrac{i}{T}) + \sum_{i=0}^{T-k-1} \sigmab'(\x'_{\frac{i}{T}},\nicefrac{i}{T}) \, (\W_{\frac{i+1}{T}} - \W_{\frac{i}{T}}) \\
    &= \x_1 - \frac{1}{T} \sum_{i=0}^{T-k-1} \overleftarrow{\f}(\x_{1 - \frac{i}{T}},1 - \nicefrac{i}{T})  + \sum_{i=0}^{T-k-1} \sigmab(\x_{1 - \frac{i}{T}},1 - \nicefrac{i}{T}) \, (\W'_{1 - \frac{i+1}{T}} - \W'_{1 - \frac{i}{T}}) \\
    &= \x_1 - \frac{1}{T} \sum_{j=k+1}\transp \overleftarrow{\f}(\x_{\frac{j}{T}}, \nicefrac{j}{T}) - \sum_{j=k+1}\transp \sigmab(\x_{\frac{j}{T}},\nicefrac{j}{T}) \, (\W'_{\frac{j}{T}} - \W'_{\frac{j-1}{T}}) \ .
\end{align}
Crucially, the terms in the two sums use the value on the \emph{right} of each interval, unlike in the usual forward SDE with the reverse-time clock $\tau$ (\cref{eq: arbitrary reverse SDE (tau)}). While for the deterministic term this is equivalent in the limit, in the stochastic term it is not.
In the limit, we get the integral equation:
\begin{align}
    \x_t = \x_1 - \int_t^1 \overleftarrow{\f}(\x_s, s) \, \diff s - \int_t^1 \sigmab(\x_s,s) \, \overleftarrow{\diff}\W'_s
\end{align}
where $\overleftarrow{\diff}$ indicates a \emph{backward Itô integral}, in the sense explained above.
The reverse-time SDE is typically written in the literature in its differential form
\begin{align}
    \diff\x_t = \overleftarrow{\f}(\x_t, t) \,\diff t+ \sigmab(\x_t,t) \, \overleftarrow{\diff}\W'_t \qquad\qquad \text{(RT SDE F-clock)}
\end{align}

\citet{anderson1982reverse} showed the stronger result that, when the forward and backward SDEs have the same time marginals \emph{and} the same exact diffusion coefficient $\sigmab(\x,t)$, \emph{the path distributions}---not just the marginals---\emph{in both directions are also identical}. %

\subsection{Negative semi-definite diffusion matrices}
\label{app: NSD diffusion matrices}

When rewriting the FPE for \cref{eq: arbitrary reverse SDE (tau)}---which is originally in the reverse-time clock $\tau$---in terms of the \emph{forward} time variable $t$, we obtain a diffusion term with a \emph{negative sign} or, equivalently, a standard diffusion term with a negative semi-definite matrix ($-\D$):
\begin{align}
    \frac{\partial p_t(\x)}{\partial t}
    = - \frac{\partial p'_\tau(\x)}{\partial \tau} 
    &= - \left( -\nabla \cdot \left[ \overrightarrow{\f'}(\x,\tau) \; p'_\tau(\x) \right] + \frac{1}{2} \nabla \cdot \left( \nabla \cdot \left[ \D'(\x, \tau) p'_\tau(\x) \right] \right) \right) \\
    &= -\nabla \cdot \left[ \overleftarrow{\f}(\x,t) \; p_t(\x) \right] - \frac{1}{2} \nabla \cdot \left( \nabla \cdot \left[ \D(\x, t) p_t(\x) \right] \right) \ .
\end{align}
This reveals a fundamental symmetry between forward- and reverse-time SDEs. Starting from the continuity equation of an ODE, adding a positive semi-definite diffusion term and adjusting the drift to preserve marginals $p_t$ results in a valid FPE corresponding to a forward-time SDE. In contrast, introducing a \emph{negative} semi-definite diffusion term and similarly adjusting the drift leads to a PDE that is not a forward-time FPE. However, under time reversal, this PDE becomes the FPE associated with a reverse-time diffusion process.

\section{The ODE and SDE are continuity equations preserving probability mass}
\label{app: continuity preserves probability}

\paragraph{Interpretation of the log density derivative.}
The integral in \cref{eq: ODE change of variable} accounts for the change of log density along a solution of the ODE.
Intuitively, \emph{the divergence of $\f$ measures the instantaneous rate of volume change under the flow}, or in other words, it describes how space contracts or dilates over time along the flow,\footnote{Note that $\f$ could still expand the volume along some directions and contract it along others. For example, if the divergence is positive, not all paths will necessarily move away from each other. Divergence merely describes the net effect.} as we briefly show next.
Consider a small volume $\mathcal{V}(t) \subseteq \R^d$ evolved by the velocity field $\f$, let $\partial \mathcal{V}(t)$ be its boundary and $\mathbf{n}(\x,t)$ the outward-pointing unit normal vector to the boundary.
Using (a) the Leibniz integral rule and (b) the divergence theorem gives:
\begin{equation}
    \frac{\diff}{\diff t} \left| \mathcal{V}(t) \right| =
    \frac{\diff}{\diff t}\int_{\mathcal{V}(t)} \diff \x
    \eqnum{\text{a}} 
    \int_{\partial\mathcal{V}(t)} \f(\x,t) \cdot \mathbf{n}(\x,t) \,\diff\mathcal{S}(\x) 
    \eqnum{\text{b}}
    \int_{\mathcal{V}(t)} \nabla \cdot \f(\x,t) \diff\x \ .
\end{equation}
If the volume is small enough that $\nabla \cdot \f(\x,t)$ is approximately constant, then $\int_{\mathcal{V}(t)} \nabla \cdot \f(\x,t) \diff\x \approx \left| \mathcal{V}(t) \right| \nabla \cdot \f(\x,t)$, and we have our result relating the divergence to local volume expansion: 
\begin{align}
    \frac{\diff}{\diff t} \log \left| \mathcal{V}(t) \right| \approx \nabla \cdot \f(\x,t) \ .
\end{align}
In addition, combining this with \cref{eq: ODE log density}, we obtain $\frac{\diff}{\diff t} \log \left| \mathcal{V}(t) \right| \approx - \frac{\diff}{\diff t} \log p_t(\x)$, suggesting that local volume and probability density are inversely proportional along the ODE flow. This is formalized below, where we show that the probability mass within a region transformed by the flow remains constant.

\paragraph{Probability conservation in the ODE.} Given a volume $\mathcal{V}(t^*) \subseteq \R^d$ at some time $t^*$ and an ODE $\frac{\diff\x}{\diff t}= \f(\x,t)$, we can define the time-dependent volume $\mathcal{V}(t)$ as the set of all points $\x_t$ obtained at time $t$ by evolving each point in $\mathcal{V}(t^*)$ under the flow of the ODE (i.e., the image of $\mathcal{V}(t^*)$ under the flow map $\phi_{t^*}^t$ from $t^*$ to $t$).
Then, the probability associated with this evolving volume remains constant over time, i.e., $\frac{\diff}{\diff t}P(\x \in \mathcal{V}(t), t) = 0$. %

To show this, we start from the ODE continuity equation:
\begin{equation}
    \frac{\partial p_t(\x)}{\partial t} = -\sum_i \frac{\partial}{\partial x_i} [f_{i}(\x,t) p(\x,t)] = - \nabla \cdot [\f(\x,t) p_t(\x)] \ .
\end{equation}
The total derivative of the probability is expressed through the Leibniz integral rule:
\begin{equation}
\label{eq:Leibniz}
\frac{\diff}{\diff t}\int_{\mathcal{V}(t)} p_t(\x)\diff\x = \int_{\mathcal{V}(t)} \frac{\partial p_t(\x)}{\partial t} \diff\x + \int_{\partial\mathcal{V}(t)} p_t(\x) %
\frac{\diff\x}{\diff t}
\cdot \mathbf{n}(\x,t) \diff \mathcal{S}(\x) \ , 
\end{equation}
where $\partial \mathcal{V}(t)$ is the boundary of $\mathcal{V}(t)$, $\mathbf{n}(\x,t)$ is the outward-pointing unit normal vector to the boundary and $\frac{\diff\x}{\diff t}
\cdot \mathbf{n}(\x,t)$ is the normal component of the velocity field on the boundary, describing how the boundary itself is moving. We can now use the divergence theorem  $\int_{\mathcal{V}} \nabla \cdot \mathbf{F}(\x) \diff\x= \int_{\partial\mathcal{V}} \mathbf{F}(\x) \cdot \mathbf{n}(\x) \diff \mathcal{S}(\x) $ to rewrite the surface term in the expression above as a volume integral:
\begin{align}
    \frac{\diff}{\diff t}\int_{\mathcal{V}(t)} p_t(\x)\diff\x  = & 
    \int_{\mathcal{V}(t)} \left( \frac{\partial p_t(\x)}{\partial t} + \nabla \cdot \left[\frac{\diff\x}{\diff t} p_t(\x)\right] \right) \diff\x \\ = & 
    \int_{\mathcal{V}(t)} \underbrace{\left( \frac{\partial p_t(\x)}{\partial t} + \nabla \cdot \left[\f(\x,t) p_t(\x)\right] \right)}_{=0 \, \mathrm{(Liouville)}} \diff\x = 0 
    \ . \nonumber
\end{align}
This fundamental result is a consequence of the Liouville equation being a continuity equation for a conserved quantity, the probability, see for example \citet{villani2009optimal}. Over time the probability density can change but the continuity equation ensures that the total probability is conserved. The Fokker--Planck equation generalizes probability conservation to the stochastic setting.

\paragraph{Probability conservation in the SDE.} We cannot apply the same analysis as for the ODE because $\mathcal{V}(t)$ is not defined in the stochastic case. Instead, we will answer the methodological question: Is there an effective deterministic velocity field $\mathbf{v}(\x,t)$ such that a volume evolving under this field conserves the probability mass of the SDE marginal $p_t(\x)$.
 
To answer this question, we apply the Leibniz integral rule \eqref{eq:Leibniz} to a volume $\mathcal{V}(t)$ moving with an unknown velocity field $\mathbf{v}(\x,t)$ while requiring total probability conservation:
\begin{equation}
    \frac{\diff}{\diff t}\int_{\mathcal{V}(t)} p_t(\x)\diff\x = \int_{\mathcal{V}(t)} \frac{\partial p_t}{\partial t} \diff\x + \int_{\partial\mathcal{V}(t)} p_t \mathbf{v} \cdot \mathbf{n} \diff \mathcal{S} = 0 \ .
\end{equation}
Using the Fokker-Planck equation and the divergence theorem, we have
\begin{align}
   \frac{\diff}{\diff t}\int_{\mathcal{V}(t)} p_t(\x)\diff\x 
    &= \int_{\partial\mathcal{V}(t)} \left( - \overrightarrow{\f}(\x,t) p_t(\x)+ \frac{1}{2} \nabla \cdot [\mathbf{D}(\x,t) p_t(\x)]   +  \mathbf{v}(\x,t) p_t(\x) \right) \cdot \mathbf{n} \diff \mathcal{S} \ .
\end{align}
For this to be zero for any volume, the integrand must be zero. Solving for $\mathbf{v}(\x,t)$, we recover the drift of an ODE with the same marginals as the SDE:
\begin{equation}
    \mathbf{v}(\x,t) = -\frac{1}{p_t(\x)} \frac{\partial p_t(\x)}{\partial t} = \overrightarrow{\f}(\x,t) - \frac{1}{2 p_t(\x)} \nabla \cdot [\mathbf{D}(\x,t) p_t(\x)]  = \f(\x,t) \ .
\end{equation}
This result shows that the Fokker--Planck is a continuity equation for the probability in the same sense as the Liouville equation. This should not surprise us because we already know that we can define both ODEs and SDEs with the same temporal evolution of the marginal.

\section{Derivations for latent SDE}

\subsection{Derivative of KL divergence}
\label{sec: derivative of kl divergence}

We will use the following shorthand and general notation in the following: $\f_t=\f(\x,t)$, $\g_t=\g(\x,t)$, $p_t=p_t(\x)$ and $q_t=q_t(\x|\y)$.
Let $p_t,q_t$ be time-varying densities that solve the continuity equation with the vector fields $\f_t,\g_t$, respectively, i.e.:
\begin{align*}
    \frac{\partial p_t}{\partial t} &= - \nabla \cdot (p_t \, \f_t) \\
    \frac{\partial q_t}{\partial t} &= - \nabla \cdot (q_t \, \g_t) 
\end{align*}
and rewrite the KL derivative in terms of the partial time derivatives of the two densities:
\begin{align}
  \frac{\diff}{\dt} \KL(q_t\|p_t)
    &= \frac{\diff}{\dt}  \int  q_t \log \frac{q_t}{p_t} \dx \nonumber \\
    &= \int \frac{\partial}{\partial t} \left[ q_t \, \log \frac{q_t}{p_t}\right] \dx \nonumber \\
    &= \int \frac{\partial q_t}{\partial t}\log \frac{q_t}{p_t}\dx +  \int q_t \frac{\partial}{\partial t}   \log q_t \dx - \int q_t \frac{\partial}{\partial t}   \log p_t \dx \nonumber \\
    &= \int \frac{\partial q_t}{\partial t}\log \frac{q_t}{p_t}\dx + \cancel{\int \frac{\partial q_t}{\partial t}  \dx} - \int \frac{q_t}{p_t} \frac{\partial p_t}{\partial t}  \dx \nonumber \ .
\end{align}
Now we can plug in the continuity equations to obtain:
\begin{align}
  \frac{\diff}{\dt} \KL(q_t\|p_t)
    &= - \int \nabla \cdot (q_t \g_t) \log \frac{q_t}{p_t}\dx + \int \frac{q_t}{p_t} \nabla \cdot (p_t \f_t) \dx \nonumber \ .
\end{align}
Integrating the first term by parts and using the divergence theorem:
\begin{align*}
    \int \nabla \cdot (q_t \g_t) \log \frac{q_t}{p_t} \, \dx
    = \cancel{\oint_{\partial\Omega} \log \frac{q_t}{p_t} \, (q_t \,\g_t \cdot \mathbf{n}) \; \diff S} - \int q_t \, \g_t \cdot \nabla \log \frac{q_t}{p_t} \, \dx \ , 
\end{align*}
where the boundary term can be understood as a flux for $\Omega$ defined as a ball with radius that tends to $\infty$, and under standard assumptions (e.g., densities decaying fast enough) it vanishes.

Analogously for the second term (once again with vanishing boundary term since the densities decay fast enough), we have:
\begin{align}
    \int \frac{q_t}{p_t} \nabla \cdot (p_t \, \f_t) \dx
    =  - \int p_t \, \f_t \cdot \nabla \frac{q_t}{p_t} \, \dx
    =  - \int q_t \, \f_t \cdot \nabla \log \frac{q_t}{p_t} \, \dx
\end{align}
and we conclude:
\begin{align}
  \frac{\diff}{\dt} \KL(q_t\|p_t)
    &= \int q_t \, \g_t \cdot \nabla \log \frac{q_t}{p_t} \, \dx - \int q_t \, \f_t \cdot \nabla \log \frac{q_t}{p_t} \, \dx \nonumber \\
    &= \E_{q_t} \big[ (\g_t - \f_t ) \cdot \nabla \log \frac{q_t}{p_t} \big] \ .
\end{align}
Going back to our case, the result above is:
\begin{align}
  \frac{\diff}{\dt} \KL(q_t\|p_t)
    &= \E_{q_t(\x \mid \y)} \big[ (\g(\x,\y,t) - \f_t(\x,t)) \cdot \nabla \log \frac{q_t(\x|\y)}{p_t(\x)} \big] \ . \label{eq: appendix KL derivative general expression}
\end{align}

\subsection{Derivative of KL divergence as forward and reverse diffusion losses}
\label{sec: derivative of kl as diffusion losses}

As shown in \cref{sec: from ODE to SDE with same marginals}, we can introduce a diffusion matrix $\D(\x,t)$ and define a forward SDE with drift $\overrightarrow{\f}(\x,t)$ and a reverse SDE with drift $\overleftarrow{\f}(\x,t)$, both with same diffusion, such that they have the same time-marginals as the ODE.
Using \cref{eq: forward SDE drift,eq: reverse SDE drift,eq: SDE drift expanded divergence}, we can write the difference between these drifts as:
\begin{align}
    \overrightarrow{\f}(\x,t) - \overleftarrow{\f}(\x,t) = \nabla \cdot \D(\x, t) + \D(\x, t) \; \nabla \log p_t(\x) 
\end{align}
and similarly for $\g(\x,\y,t)$. In the following we again apply the shorthand from the previous section.
The score difference can thus be written in terms of forward and reverse drifts:
\begin{equation}
    \D_t \; \nabla \log \frac{q_t}{p_t} = \big(\overrightarrow{\g_t} -\overleftarrow{\g_t}\big) - \big(\overrightarrow{\f_t} - \overleftarrow{\f_t}\big)
\end{equation}
and assuming that the diffusion matrix $\D_t = \sigmab_t\transp\sigmab_t$ is invertible:
\begin{align}
    \nabla \log \frac{q_t}{p_t} = \D_t ^{-1} \left[\big(\overrightarrow{\g_t} -\overleftarrow{\g_t}\big) - \big(\overrightarrow{\f_t} - \overleftarrow{\f_t}\big) \right] \ .
\end{align}
From the results in \cref{sec: from ODE to SDE with same marginals} we note that:
\begin{align}
    \f_t &= \tfrac12 (\overrightarrow{\f_t} + \overleftarrow{\f_t}) \\
    \g_t &= \tfrac12(\overrightarrow{\g_t} + \overleftarrow{\g_t}) \ .
\end{align}
Using the facts above, we can rewrite \cref{eq: appendix KL derivative general expression} as follows:
\begin{align}
    \frac{\diff}{\diff t}\KL\big(q_t \,\|\, p_t\big)
    &= \E_{q_t} \Big[ 
        \big(\g_t - \f_t \big)\transp \nabla\log \frac{q_t}{p_t}
    \Big] \\
    &= \E_{q_t}
    \left[
        \big(\g_t-\f_t\big)\transp  \D_t^{-1} \Big((\overrightarrow{\g_t}-\overleftarrow{\g_t})-(\overrightarrow{\f_t} - \overleftarrow{\f_t})\Big)
    \right] \\
    &= \mathbb{E}_{q_t}
    \left[
        \Big(\tfrac12(\overrightarrow{\g_t}+\overleftarrow{\g_t})-\tfrac12(\overrightarrow{\f_t}+\overleftarrow{\f_t})\Big)\transp  \D_t^{-1} \Big((\overrightarrow{\g_t}-\overleftarrow{\g_t})-(\overrightarrow{\f_t} - \overleftarrow{\f_t})\Big)
    \right] \\
    &= \frac12 \, \E_{q_t}
    \Big[
        \big((\overrightarrow{\g_t} - \overrightarrow{\f_t})+(\overleftarrow{\g_t} - \overleftarrow{\f_t})\big)\transp \D_t^{-1} \big((\overrightarrow{\g_t} - \overrightarrow{\f_t}) - (\overleftarrow{\g_t} - \overleftarrow{\f_t})\big)
    \Big] \\
    &= \frac12 \, \E_{q_t}
    \Big[
    (\overrightarrow{\g_t}-\overrightarrow{\f_t})\transp \D_t^{-1}(\overrightarrow{\g_t}-\overrightarrow{\f_t}) - (\overleftarrow{\g_t}-\overleftarrow{\f_t})\transp \D_t^{-1}(\overleftarrow{\g_t}-\overleftarrow{\f_t}) \ , 
    \Big] 
\end{align}
where the last step follows by expanding the product and using the fact that $\D^{-1}$ is symmetric.
Defining:
\begin{align}
    \mathcal{D}(\f,\g,\y) &\coloneqq \frac{1}{2}
    \int_0^1 \ \E_{q_t(\x\mid\y)}\left[ \left(\f(\x,t)- \g(\x,\y,t)\right)\transp \, \D(\x,t)^{-1} \, \left(\f(\x,t)- \g(\x,\y,t)\right) \right] \diff t 
\end{align}
we can rewrite that as 
\begin{align}
    \int_0^1 \frac{\diff}{\diff t}\KL\big(q_t \,\|\, p_t\big) \,\dt
    &= \mathcal{D}(\overrightarrow{\f},\overrightarrow{\g},\y)   -  \mathcal{D}(\overleftarrow{\f},\overleftarrow{\g},\y) \ .
\end{align}

\section{SDE ELBO via KL divergence between path measures}
\label{app: latent SDE path}

Here we provide an alternative derivation of the SDE ELBO, i.e., the ELBO arising from having a latent process that follows an SDE. As discussed, in this case $p_1(\x_1) = \int p(\x_1|\x_0) p(\x_0)\diff \x_0$ cannot be computed directly via a change of variables and is intractable in most cases. The KL divergence $\KL(q_1(\x_1|\y)\|p_1(\x_1))$ in the ELBO \eqref{eq: general ELBO with gap} is therefore intractable.
In \cref{sec: latent SDE} we addressed this by upper-bounding $\KL(q_1(\x_1|\y)\|p_1(\x_1))$ with a tractable expression, to obtain a looser log-likelihood bound $\SDEELBO(\y) \leq \ODEELBO(\y) \leq \log p(\y)$.
Here, we will derive exactly the same bound by using properties of KL divergences over path measures.

\paragraph{Path distributions for generative and variational models.} We start by specifying the path distribution $p(\X)$ of the generative model over \emph{latent paths} $\X = \{\x_t\}_{t \in [0,1]}$ rather than over the final-time variable $\x_1$ only. The path distribution is defined through the SDE \eqref{eq:SDE} (where the right arrow is omitted in the following for simplicity): 
\begin{align}
    \diff\x_t & = \f(\x_t,t)\diff t+ \sigmab(\x_t,t) \diff\W_t \ , \qquad \x_0 \sim p_0(\x_0) \ .
\end{align}
Similarly, the variational distribution $q(\X|\y)$ is defined through a second SDE:
\begin{align}
    \diff\x_t & = \g(\x_t,\y,t)\diff t+ \sigmab(\x_t,t) \diff\W_t \ , \qquad \x_0 \sim q_0(\x_0|\y) %
    \label{eq: variational SDE}
\end{align}
that has a data-dependent drift $\g(\x_t,\y,t)$ and shares the diffusion coefficient with the generative SDE. The latter is a necessary condition to get a finite KL divergence in \cref{eq:ELBO_SDE_definition} \citep{archambeau2007gaussian,nielsen2023diffenc,sarkka2019applied}.
 
As discussed in \cref{sec: Generative models from differential equations}, while there is no standard definition of densities over paths in the usual sense, we will still informally use them to keep notation simple and more similar to standard probabilistic modeling.
All the results presented here can be made rigorous by using path measures---for example, $q(\X|\y)$ corresponds to a path measure $\Qpath(\Omega | \y)$ which assigns a probability to a set of paths $\Omega \subseteq C([0,1], \R^d)$ conditional on $\y$.

\paragraph{SDE ELBO.}
First we show that the path KL divergence $\KL(q(\X|\y)||p(\X))$ is an upper bound to marginal KL divergence $\KL(q_1(\x_1|\y)||p_1(\x_1))$.
Denoting by $\X_{<1}$ the path excluding $t=1$, by the chain rule of the KL divergence we have:\footnote{See \citet[Theorem~2.4]{leonard2014some} for a rigorous result on path measures.}
\begin{align}
    \KL(q(\X|\y)||p(\X))
    = \KL(q_1(\x_1|\y)||p_1(\x_1)) + \E_{q_1(\x_1|\y)}[\underbrace{\KL(q(\X_{<1}|\x_1,\y)||p(\X_{<1}|\x_1)}_{\geq 0}] \ .
\end{align}
Using this relationship, we can now define the SDE ELBO by bounding the ODE ELBO \eqref{eq: ODE ELBO definition}:
\begin{align}
    \ODEELBO(\y)
    &= \E_{q_1(\x_1|\y)} \left[ \log p(\y | \x_1)\right] - \KL(q_1(\x_1|\y) \| p_1(\x_1)) \nonumber \\
    & \label{eq:ELBO_SDE_definition}
    \geq \E_{q_1(\x_1|\y)} \left[ \log p(\y | \x_1)\right] - \KL(q(\X|\y)||p(\X)) \eqqcolon \SDEELBO(\y) \ .
\end{align}
In \cref{app: girsanov} we make an informal derivation of the path KL divergence that we can insert in \eqref{eq:ELBO_SDE_definition} to arrive at:
\begin{align}
    \label{eq:ELBO_SDE appendix}
    \SDEELBO(\y) &=   \underbrace{\E_{q_1(\x_1|\y)}\left[ \log p(\y|\x_1) \right]}_{\text{Reconstruction}} - \underbrace{ \KL(q_0(\x_0|\y) \| p_0(\x_0))}_{\text{Prior}}  - \underbrace{\mathcal{D}(\f,\g,\y)}_{\text{Diffusion}} \\
     \label{eq:diffusion}
    \mathcal{D}(\f,\g,\y) &= \frac{1}{2}
     \int_0^1 \E_{q_t(\x|\y)}\left[ \left(\f(\x,t)- \g(\x,\y,t)\right)\transp \D(\x,t)^{-1} \left(\f(\x,t)- \g(\x,\y,t)\right) \right]\diff t\ .
\end{align}
The diffusion coefficient is often chosen to be a scalar function $\sigmab(\x,t) = \sigma(\x,t) \I$. Then, $\D(\x,t)=\sigmab(\x,t)\sigmab\transp (\x,t)= \sigma^2(\x,t) \I$ and the diffusion term simplifies to:
\begin{align}
    \label{eq:diffusion_simple}
    \mathcal{D}(\f,\g,\y) &= \frac{1}{2}
     \int_0^1 \E_{q_t(\x|\y)}\left[ \frac{\left\|\f(\x,t)- \g(\x,\y,t)\right\|^2}{\sigma^2(\x,t)}  \right]\diff t\ .
\end{align}

\subsection{Informal derivation of the path KL divergence}\label{app: girsanov}

The KL divergence between path measures in \eqref{eq:ELBO_SDE_definition}, necessary to compute the ELBO for the SDE, can be directly derived in continuous time using the Girsanov theorem \citep{huang2021variational,song2021maximum,sarkka2019applied}. Here we will instead derive it using the Euler--Maruyama discretization of the SDE starting from finite $\Delta t$ and taking the infinitesimal limit, as also done, e.g., by \citet{archambeau2007gaussian} and \citet{bartosh2024neural}.\footnote{Other works \citep[e.g.,][]{ho2020denoising,kingma2021variational,nielsen2023diffenc} employ a different discretization that has the same continuous-time limit---see, for example, \citet[][Appendix~E]{song2020score} for a discussion.} 
This discretization approach is close in spirit to an informal derivation of the Girsanov theorem \citep{sarkka2019applied}.

We consider the two discrete-time processes with $T$ steps and $T+1$ $t$-values: 
corresponding to the generative model $p$ and the inference model $q$:
\begin{align}
     p:\mbox{\hspace{0.5cm}} \Delta \x_l & = \x_{t_{l+1}} - \x_{t_{l}} = \f_{l} \Delta t + \sigmab_l \Delta \W_l \\
     q:\mbox{\hspace{0.5cm}} \Delta \x_l & = \x_{t_{l+1}} - \x_{t_{l}} = \g_{l} \Delta t + \sigmab_l \Delta \W'_l \ ,
\end{align}
where $\Delta \W_l$ and $\Delta \W'_l$ are zero-mean Gaussians with variance $\Delta t = \nicefrac{1}{T}$, and we use the shorthand $\f_{l}= \f(\x_{t_{l}},t_l)$ and similarly for $\g_l$, $\sigmab_l$ and $\D_l = \sigmab_l \sigmab_l\transp$.
Assuming $t_l = l \, \Delta t$, these two processes can be seen as discrete-time analogues of the inference and generative SDEs considered earlier.
The joint distributions over $\X=\x_{t_0},\ldots,\x_{t_T}$ are:
\begin{align}
    p(\X) & = p_0(\x_{0}) \prod_{l=0}^{T-1} p_{t_{l}}(\x_{t_{l+1}}|\x_{t_{l}}) = p_0(\x_{0}) \prod_{l=0}^{T-1} \mathcal{N}(\x_{t_{l+1}}|\x_{t_{l}}+\f_{l}\Delta t,\sigmab_l \sigmab_l\transp \Delta t) \\
     q(\X|\y) & = q_0(\x_0|\y) \prod_{l=0}^{T-1} q_{t_{l}}(\x_{t_{l+1}}|\x_{t_{l}},\y) = q_0(\x_0|\y) \prod_{l=0}^{T-1} \mathcal{N}(\x_{t_{l+1}}|\x_{t_{l}}+\g_{l}\Delta t,\sigmab_l \sigmab_l\transp \Delta t) \ ,
\end{align}
where we have left the prior distributions unspecified for now. We plug in these distributions into the ELBO Equation (\ref{eq:ELBO_SDE_definition}). 
The KL divergence is the expectation with respect to $q(\X|\y)$ of the following log-likelihood ratio:
\begin{align}
    \label{eq:log_likelihood_ratio}
   \log \frac{p(\X)}{q(\X|\y)} = \log \frac{ p_0(\x_{0}) }{ q_0(\x_0|\y) } - \frac{1}{2\Delta t} \sum_{l=0}^{T-1}    & \left\{ (\Delta \x_l - \f_{l} \Delta t)\transp \D^{-1}_l (\Delta \x_l - \f_{l} \Delta t)- \right.\\& \, \, \left. (\Delta \x_l - \g_{l} \Delta t)\transp \D^{-1}_l (\Delta \x_l - \g_{l} \Delta t)  \right\} \ . \nonumber 
\end{align}
We can rewrite the summation term in a form that makes it more suitable for taking expectation and for the $T\to\infty$ limit:
\begin{align}
    \label{eq:path_lntegral}
    \underbrace{- \frac{\Delta t}{2}\sum_{l=0}^{T-1} (\f_{l}-\g_{l})\transp \D^{-1}_l (\f_{l}-\g_{l}) }_{\text{(i)}} + \underbrace{\sum_{l=0}^{T-1} (\f_{l}-\g_{l})\transp \D^{-1}_l (\Delta \x_l - \g_{l} \Delta t)}_{\text{(ii)}}   \ .
\end{align}
The expectation of (ii) with respect to $q(\X|\y)$ is zero:
\begin{align}
    \E_{q(\X|\y)}\left[ \sum_{l=0}^{T-1} (\f_{l}-\g_{l})\transp \D^{-1}_l (\Delta \x_l - \g_{l} \Delta t) \right]
    &= \sum_{l=0}^{T-1} \E_{q_l} \Bigg[ (\f_{l}-\g_{l})\transp \D^{-1}_l  \ \underbrace{\E_{q_{l+1|l}} \left[ \x_{l+1} - \x_l - \g_{l} \Delta t\right]}_{=0} \Bigg] \ ,
\end{align}
where we used the shorthand $q_l = q_{t_l}(\x_{t_{l}}|\y)$ and $q_{l+1|l} = q_{t_{l+1}}(\x_{t_{l+1}}|\x_{t_{l}},\y)$.
We can then take the limit for $T \to \infty$ of the expectation of (i):
\begin{align}
    \E_{q(\X|\y)} & \left[- \frac{\Delta t}{2}\sum_{l=0}^{T-1} (\f_{l}-\g_{l})\transp \D^{-1}_l (\f_{l}-\g_{l}) \right] 
    = - \frac{1}{2}\sum_{l=0}^{T-1} \mathbb{E}_{q_l} \left[ (\f_{l}-\g_{l})\transp \D^{-1}_l (\f_{l}-\g_{l}) \right] \Delta t \\
    &\xrightarrow[]{T \to \infty} \underbrace{- \frac{1}{2}\int_0^1 \E_{q_t(\x|\y)}\left[  (\f(\x,t)-\g(\x,\y,t))\transp \D^{-1}_l(\f(\x,t)-\g(\x,\y,t)) \right] d t}_{\mathrm{Diffusion}}\ . \nonumber
    \label{eq: diffusion loss}
\end{align}
We have now calculated the diffusion term in the ELBO \eqref{eq:ELBO_SDE_definition}. The remaining term appearing in the log-likelihood ratio \eqref{eq:log_likelihood_ratio} is the prior term: 
\begin{equation}
     \E_{q(\X|\y)} \left[ \log \frac{ p_0(\x_{0}) }{ q_0(\x_0|\y) } \right] = \E_{q_0(\x|y)} \left[ \log \frac{ p_0(\x) }{ q_0(\x|\y) } \right] = \underbrace{- \KL(q_0(\x|\y),p_0(\x))}_{\mathrm{Prior}} \ .  
\end{equation}
We can now write the ELBO for the SDE \eqref{eq:ELBO_SDE_definition} in the continuous limit as stated in \cref{eq:ELBO_SDE}.

\section{Connection to standard derivations of score and flow matching}\label{app:score and flow}

In this appendix, we elaborate on the derivation of the score and flow matching objectives presented in \cref{sec: score and flow matching} and draw connections to alternative derivations of the same objectives found in the literature.

\subsection{Score matching}

Here we will show that the optimal score function is indeed the marginal score, i.e.,
\begin{align}
    \s^*(\x,t) = \argmin_\s \E_{q(\y)} \left[ \mathcal{D}_{\mathrm{Score}} \right] = \nabla \log q_t(\x) \ ,
\end{align}
where $q(\y)$ denotes the (unknown) data distribution. To derive this result, we average Equation \cref{eq: diff loss as denoising score matching} over $q(\y)$: 
\begin{align}
    \E_{q(\y)} \left[ \mathcal{D}_{\mathrm{Score}} \right] 
    & = \frac{1}{2} \int_0^1 \E_{q_t(\x|\y)q(\y)}\left[ \sigma^2(\x,t) \left(  ||\s(\x, t)||^2 - 2 \s\transp (\x, t) \nabla \log q_t(\x| \y)  +   ||\nabla \log q_t(\x|\y)||^2 \right)  \right]  \diff t\ . 
    \label{eq: diff loss denoising score matching expectation q(y)}
\end{align}
We consider each term in turn. The first term does not depend on $\y$, thus:
\begin{align*}
    \E_{q_t(\x|\y)q(\y)} \left[ \sigma^2(\x,t) || \s(\x, t)||^2 \right] 
    = \E_{q_t(\x)} \left[ \sigma^2(\x,t) || s(\x, t)||^2  \right]  \ .
\end{align*}
The second term can be rewritten as follows:
\begin{align*}
     \E_{q_t(\x|\y)q(\y)}\left[ \sigma^2(\x,t) \s\transp (\x, t) \nabla \log q_t(\x| \y) \right]
     &= \E_{q_t(\x)}\left[ \sigma^2(\x,t)  \s\transp (\x, t) \  \E_{q_t(\y|\x)}\left[\nabla \log q_t(\x| \y)  \right]\right] \\
     &= \E_{q_t(\x)}\left[ \sigma^2(\x,t) \s\transp (\x, t) \ \nabla \log q_t(\x)\right] \ , 
\end{align*}
where we used Bayes $q_t(\y|\x)=\frac{q_t(\x|\y)q(\x)}{q(\y)}$ and
\begin{align*}
    \E_{q_t(\y|\x)}\left[\nabla \log q_t(\x|\y)\right]
    &= \E_{q_t(\y|\x)}\left[\nabla \log q_t(\y|\x)\right] + \nabla \log q_t(\x) \\
    &= \int \nabla q_t(\y|\x)  d\y + \nabla \log q_t(\x) \\
    &= \nabla \underbrace{ \int  q_t(\y|\x) d\y}_{=1} + \nabla \log q_t(\x) 
\end{align*}
to write the expression in terms of the marginal score $\nabla \log q_t(\x)$.
\cref{eq: diff loss denoising score matching expectation q(y)} can therefore be rewritten as:
\begin{align*}
    \E_{q(\y)} \left[ \mathcal{D}_{\mathrm{Score}} \right] 
    &= \frac{1}{2} \int_0^1 \E_{q_t(\x)}\left[ \sigma^2(\x,t) \left(  ||\s(\x, t)||^2 - 2 \s\transp (\x, t) \nabla \log q_t(\x)  \right)  \right] \diff t\\
    & \qquad +\frac{1}{2} \int_0^1 \E_{q_t(\x|\y)q(\y)}\left[ \sigma^2(\x,t) || \nabla \log q_t(\x| \y)||^2  \right]\diff t\\
    &= \frac{1}{2} \int_0^1 \E_{q_t(\x)} \left[ \sigma^2(\x,t) \left(  \left\| \s(\x, t)- \nabla \log q_t(\x)\right\|^2 - || \nabla \log q_t(\x)||^2 \right) \right]\diff t \\ 
    & \qquad +\frac{1}{2} \int_0^1 
    \E_{q_t(\x|\y)q(\y)}\left[ \sigma^2(\x,t) ||  \nabla \log q_t(\x | \y) ||^2 \right] \diff t\\
    &= \frac{1}{2} \int_0^1 \E_{q_t(\x)} \left[ \sigma^2(\x,t) \left(  \left\| \s(\x, t)- \nabla \log q_t(\x)\right\|^2  \right) \right]\diff t+ \mathrm{const.} \ ,
\end{align*}
where we completed the square in the first integral and in the last expression arrived at the sought objective with the constant collecting terms independent of $\s(\x,t)$.

This shows that the expectation of the diffusion loss \cref{eq: diff loss as denoising score matching} over the data distribution defines a least squares objective with optimum $\s^*(\x,t) = \nabla \log q_t(\x)$. So in the asymptotic limit of infinite data, the optimization of the diffusion term leads to learning of the score $\nabla \log q_t(\x)$. 

In the diffusion models literature (see, e.g., \citet{song2020score}), it is common to derive the score-matching formulation by proceeding in the reverse order. Typically, one starts from a \emph{marginal} score matching objective like the following:
\begin{align*}
    \frac{1}{2} \int_0^1 \E_{q_t(\x)}\left[ \lambda(\x,t) \left\| \s(\x, t)- \nabla \log q_t(\x)\right\|^2 \right]\diff t\ ,
\end{align*}
which is a weighted average over time, with (almost) arbitrary weights $\lambda(\x,t)$. While this objective is intractable, as it requires access to the true marginal score, it has the same optimum $\s^*(\x,t)$ as the expectation of \cref{eq: diff loss as denoising score matching}, which can be optimized in practice. However, unless $\lambda(\x,t)$ is exactly $\sigma^2(\x,t)$, as we found above, this is not linked with maximum likelihood (this connection was discussed, e.g., by \citet{song2021maximum}, \citet{huang2021variational}, and \citet{kingma2021variational}).

\subsection{Flow matching}

To connect to the flow matching literature \citep{lipman2022flow,liu2022flow}, we calculate the expectation of the expression above in the same way as done for the score objective: 
\begin{align}
    \E_{q(\y)}\left[\mathcal{D}_{\mathrm{Flow}} \right]
    = & 2\int_0^1  \E_{q_t(\x)}\left[\frac{\left\| \v(\x,t) - \g(\x,t) \right\|^2 - ||\g(\x,t)||^2}{\sigma^2(\x,t)}  \right]\diff t \ + \\ & 2\int_0^1  \E_{q_t(\x|\y)q(\y)}\left[\frac{||\g(\x,\y,t)||^2}{\sigma^2(\x,t)} \right]\diff t \ ,
\end{align}
where we have introduced the ``marginalized'' drift:
\begin{equation}
    \label{eq:g0 marginal}
    \g(\x,t) = \frac{1}{q_t(\x)} \int \g(\x,\y,t) q_t(\x|\y) q(\y) d\y \ . 
\end{equation}
So in the limit of infinite data, we get a least squares objective with optimum $\v^*(\x,t) = \g(\x,t)$. 

\paragraph{Connection between conditional and marginal Liouville equations.} We can get further insight about this drift by relating the conditional Liouville equation: $\frac{\partial q_t(\x|\y)}{\partial t} = - \nabla \cdot \left[\g(\x,\y,t) q_t(\x|\y)\right]$ to the Liouville equation for the marginal $\frac{\partial q_t(\x)}{\partial t} = - \nabla\cdot \left[\tilde{\g}(\x,t) q_t(\x)\right]$:  
\begin{align}
    \frac{\partial}{\partial t} q_t(\x) & = \frac{\partial}{\partial t} \int q_t(\x|\y) q(\y) d\y =  \int \frac{\partial q_t(\x|\y)}{\partial t} q(\y) d\y = 
    - \int \nabla \cdot \left[\g(\x,\y,t) q_t(\x|\y)\right] q(\y) d\y \\ 
    & = - \nabla \cdot \underbrace{\int \g(\x,\y,t) q_t(\x|\y) q(\y) d\y}_{\tilde{\g}(\x,t) q_t(\x)}  \nonumber \ .
\end{align}
This indeed confirms that the drift $\g(\x,t)$ Equation (\ref{eq:g0 marginal}) is the drift $\tilde{\g}(\x,t)$ in the Liouville equation for the marginal $q_t(\x)$.

\end{document}